\documentclass[10pt,twocolumn,twoside]{IEEEtran}
\usepackage{amsmath,amsfonts}
\usepackage{algorithm2e}
\usepackage{array}
\usepackage[caption=false,font=normalsize,labelfont=sf,textfont=sf]{subfig}
\usepackage{textcomp}
\usepackage{stfloats}
\usepackage{url}
\usepackage{verbatim}
\usepackage{graphicx}
\usepackage{breqn}
\usepackage{bbm}
\usepackage{cite}
\usepackage{arydshln}

\newcommand\revision[1]{\textcolor{black}{#1}}
\usepackage{tabularray}
\usepackage{multirow,booktabs}
\usepackage[table,xcdraw]{xcolor}
\usepackage[normalem]{ulem}
\useunder{\uline}{\ul}{}

\DeclareMathOperator*{\argmin}{arg\,min}
\begin{document}

\title{Gradient-free Post-hoc Explainability Using Distillation Aided Learnable Approach}

\author{Debarpan Bhattacharya,~\IEEEmembership{ Student Member,~IEEE,} Amir H. Poorjam, Deepak Mittal, \\ and Sriram Ganapathy,~\IEEEmembership{Senior Member,~IEEE}
\thanks{D. Bhattacharya is a PhD student at the Electrical Engineering department, Indian Institute of Science, Bangalore, India, 560012.
A. Poorjam\textsuperscript{\dag} is a Data Science Specialist in Global Clinical Department at Lundbeck A/S in Copenhagen, Denmark.
D. Mittal\textsuperscript{\dag} is a PhD student at the Artificial Intelligence department, Indian Institute of Technology, Hyderabad.
S. Ganapathy is an Associate Professor and the Director of the Learning and Extraction of Acoustic Pattern (LEAP) laboratory, Electrical Engineering, Indian Institute of Science, Bangalore, India.\\
\textsuperscript{\dag}This work was done during employment at Verisk.}
\thanks{Manuscript submitted on July 7, 2023.}}

\markboth{Journal of \LaTeX\ Class Files,~Vol.~XX, No.~YY, August~2023}%
{Bhattacharya \MakeLowercase{\textit{et al.}}: D\MakeLowercase{istillation} A\MakeLowercase{ided} \MakeLowercase{e}X\MakeLowercase{plainability} (DAX) \MakeLowercase{framework for} XAI}

\maketitle
\begin{abstract}
The recent advancements in artificial intelligence (AI), with  the release of several large models having only query access, make a strong case for explainability of deep models in a post-hoc gradient free manner. In this paper, we propose a framework, named distillation aided explainability (DAX), that attempts to generate a saliency-based explanation in a model agnostic gradient free application. The DAX approach poses the problem of explanation in a learnable setting with a mask generation network and a distillation network. The   mask generation network learns to generate the multiplier mask that finds the salient regions of the input, while  the student distillation network aims to approximate the local behavior of the  black-box model. We propose a joint optimization of the two networks in the DAX framework using the locally perturbed input samples, with the targets derived from input-output access to the black-box model.  
\revision{We extensively evaluate DAX across different modalities (image and audio), in a classification setting,  using a diverse set of evaluations (intersection over union with ground truth, deletion based and subjective human evaluation based measures) and benchmark it with respect to $9$ different methods.  In these evaluations, the DAX significantly outperforms the existing approaches on all modalities and evaluation metrics.
}
\end{abstract}

\section{Introduction}
In today's machine learning, deep neural models are the mostly widely used manifestation of mapping functions from the feature space to the target labels. 
Current deep models with state-of-the-art capabilities consist of specialized units for processing different kinds of modalities, for example, fully connected neural networks~\cite{lecun2015deep}, long short-term memory units~\cite{hochreiter1997long}, attention units~\cite{BahdanauCB14},  transformers~\cite{vaswani2017attention}, auto encoders~\cite{rumelhart1985learning} etc. 
Diverse combinations of such architectures at massive scale have manifested in the development of foundation models, with astounding performance on several tasks, for example, Vision Transformer (ViT)~\cite{dosovitskiy2020image} in computer vision, Whisper~\cite{radford2022robust} in speech processing and GPT-4~\cite{openai2023gpt4} in natural language processing. 
In several domains,  the sequence-to-sequence models based on   transformer architecture \cite{vaswani2017attention} have achieved 
state-of-art results and in some cases elicited super-human performances. 

While the models may have illustrated impressive performance, the deep complex architectures are inherently non-explainable. 
\revision{While knowledge distillation can provide simpler modeling  solutions \cite{gou2023multi, gou2022multilevel},} the questions about data contamination and leakage  \cite{kaufman2012leakage}, as well as the over-estimation of the model performance \cite{patel2008investigating}, continue to be pertinent in this age of large language models (LLMs) \cite{chatgpt}. 
The research direction of explainable artificial intelligence (XAI) attempts to provide solutions for this scenario.

Explaianability of deep models is a crucial requirement if they are deployed in safety-critical scenarios~\cite{doshi2017accountability, lipton2018mythos, holzinger2017we, caruana2015intelligible}. For example, in domains like autonomous driving, finance, and healthcare, lack of explainability in the complex neural models may turn the practitioners away from exploiting the impeccable results that these models may offer. Further, XAI approaches can also help in identifying spurious correlations~\cite{liao2020questioning, chouldechova2020snapshot} and biases~\cite{ras2022explainable, wang2015causal} present in the datasets and models. 

\subsection{Types of XAI methods}
The explainability problem first assumes a framework where the model (referred to as the black-box in our discussion) configuration, architecture information, and downstream tasks, are given. 
The setting explored in this paper is one in which an explanation is sought for an already trained black-box model with no further provision to modify/retrain it,  called the \textit{post-hoc explainability} setting. Based on different criteria, post-hoc methods can be further  categorised based on,

\noindent\textbf{Model architecture}: The XAI approaches can be, (a) \textit{Model-specific}: when the XAI method works only for a specific black-box architecture (for example, XAI for
CNN-based models~\cite{zhou2016learning, chattopadhay2018grad}); or (b) \textit{Model-agnostic}: when the XAI method is agnostic to the architecture of the black-box~\cite{DBLP:journals/corr/SmilkovTKVW17, sundararajan2017axiomatic, ribeiro2016should}.

\noindent\textbf{Access to black-box}: Depending on the access requirements of the XAI methods, they can be classified as, (a) \textit{gradient-based approaches}: XAI methods using gradient access to black-box like~\cite{smilkov2017smoothgrad, sundararajan2017axiomatic, chattopadhay2018grad}; or (b) \textit{gradient-free approaches}: XAI methods that only need input-output access without gradient access like, perturbation-based methods~\cite{zeiler2014visualizing, ribeiro2016should, petsiuk2018rise}. 

\noindent\textbf{Locality of explanations}: As the black-box is highly nonlinear, the explanation can be, (a) \textit{local} - the methods that attempt to generate explanations for each test example like \cite{selvaraju2017grad, petsiuk2018rise}, or (b) \textit{global} - the methods that attempt to form a complete explanation of the model function, like the works reported in \cite{natesan2020model, pedapati2020learning}.  
\begin{figure*}[t!]
  \centering
  \includegraphics[width=0.78\textwidth, height=0.38\textwidth]{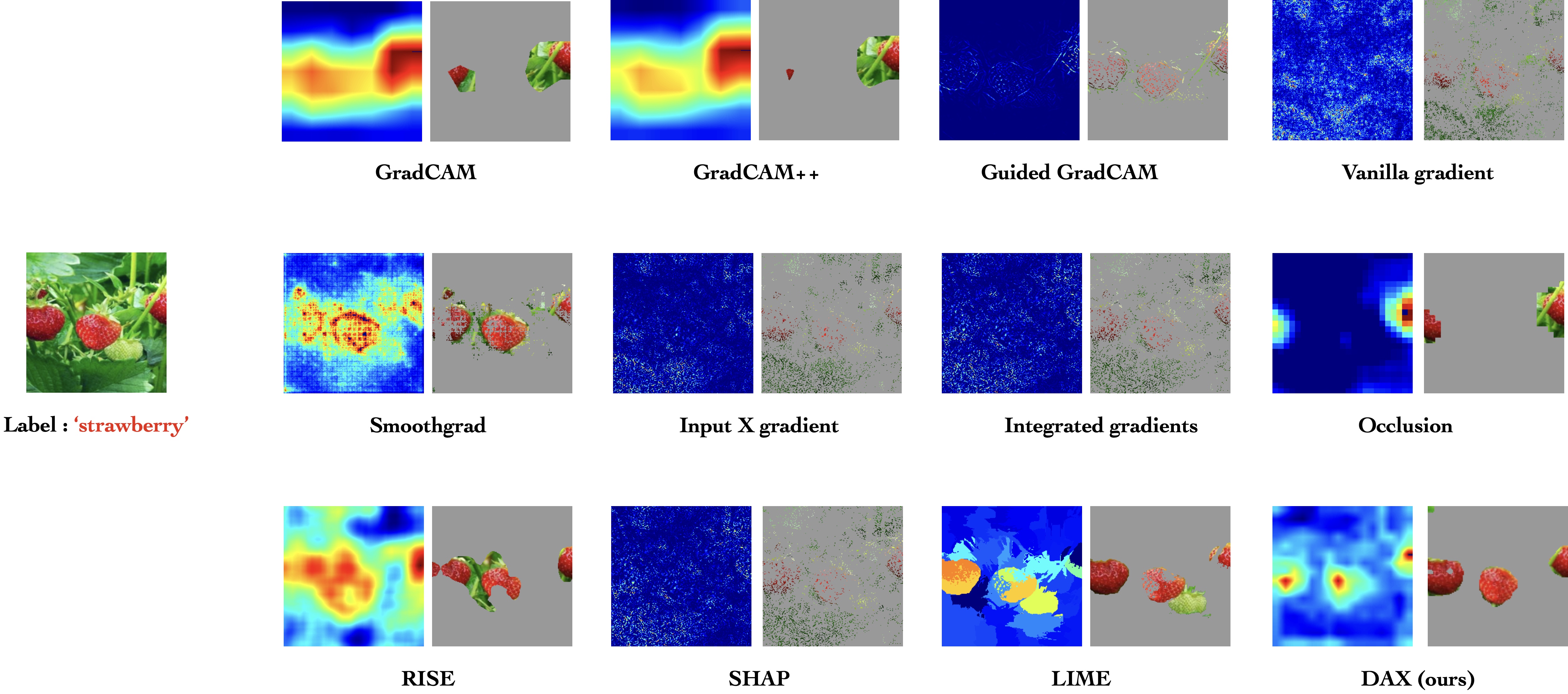}
  \caption{Comparing various XAI methods and DAX for input \textit{strawberry} image with the explanation heat map and the mask multiplied input.  
  }
  \vspace{-0.1in}
  \label{intro-saliency-examples}
\end{figure*}

With the increasing number of large models being released having  only query access (e.g. ChatGPT~\cite{chatgpt}, DALL-E~\cite{ramesh2021zero}), developing post-hoc gradient-free XAI approaches   is crucial.


\subsection{Contributions from the proposed framework}

In computer vision and audio based models, the input dimensionality is  high, $\mathcal{O}$($10^4$) or more. (for example,  high-resolution images of size $224$ X $224$ result in input dimension of $4e^4$). To avoid dealing with such high dimensional locality, the prior works  compute image segments and consider each of them as a dimension rather than the raw pixels. With this approach, the input dimensionality is reduced to $\mathcal{O}$($10^1$). However, explanations generated this way can be \textit{unreliable}, \textit{erroneous} and \textit{imprecise} as discussed in detail in Section~\ref{sec:post-hoc-methods}. 

We propose a novel \textit{learnable}, distillation-based approach to post-hoc gradient-free explainability that uses two  non-linear networks, a) mask-generation network and, b) student network to compute explanations. The mask-generation network selects salient parts of the local input. With the mask-multiplied input, the student network attempts to locally approximate the black-box predictions. Using a joint training framework with perturbation samples, the mask-generation network finds the salient region of the input image as the explanation. 

The proposed framework, termed as distillation aided explanations (DAX), is shown to compute explanations that are better than existing approaches in various aspects, although it  operates directly on the $\mathcal{O}$($10^4$) dimensional input space of the black-box, 
while adding a minimal computational overhead. 

This paper extends our prior work using student-teacher distillation~\cite{bhattacharya-dame-nips23}.
The following are the major contributions in this paper,  
\begin{enumerate}
    \item Mathematical formulation of the DAX framework with the optimization.
    \item A simplified approach which generates a single explanation mask over all the input channels.
    \item Detailed analysis on the sensitivity of the approach as well as with samples that are incorrectly predicted by the black-box model.
    \item Extensive comparison with $11$ baseline systems on image tasks with ViT/ResNet black-box models and  audio tasks with ResNet/LSTM based black-box models.
    \item Improved performance on the IoU metric,  \revision{deletion area under the curve (AUC) and subjective evaluations} for both the audio and image tasks compared to all the baselines systems. 
\end{enumerate}
\begin{figure*}[t!]
    \centering
    \includegraphics[width=18cm, height=6.3cm]{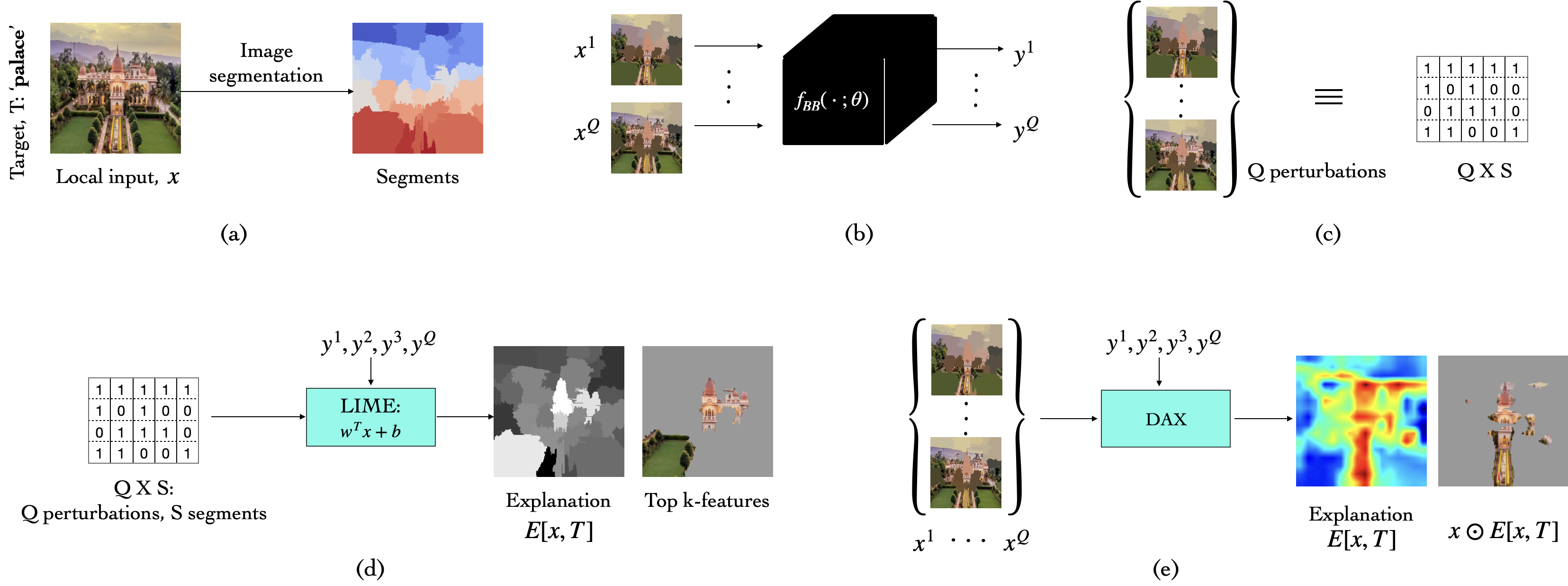}
    \caption{
    Contrasting the local linear approximation approach LIME\cite{ribeiro2016should} for explainability and DAX (this work). (a)  LIME segments the input image $x$. Let,   the number of segments be $S$. (b) Then, it masks off segments randomly and generates the corresponding black-box responses, $y^1, ..., y^Q$ where $Q$ is the number of perturbation samples. (c) Each of the perturbed image is represented by a binary row of size $S \times 1$ where a $1$ represents masking off the corresponding segment. (d) Using the binary matrix of size, $Q \times S$  as input, and $y^1, ..., y^Q$ as targets, the LIME fits a linear model. Following the training, the linear weights denote the explanation of the input, in terms of weights corresponding to the segment locations. (e) DAX, in contrast,  operates on the image space directly and uses non-linear local approximation to generate the explanation. }
    \vspace{-0.1in}
    \label{fig:lime-working-principle}
\end{figure*}

\section{Related prior work}
\subsection{Post-hoc explainability approaches}
\label{sec:post-hoc-methods}
An illustration of the explainability provided by various methods for an image example is shown in Figure~\ref{intro-saliency-examples}.  
There are several prior works using gradient access based explainability, namely,
\begin{enumerate}
    \item Vanilla gradient \cite{simonyan2014visualising}, gradient$\times$input \cite{shrikumar2016not, shrikumar2017learning}, smoothgrad~\cite{smilkov2017smoothgrad} and Integrated gradients \cite{yosinski2021understanding}.\item Class activation maps (CAM) \cite{zhou2016learning}, Gradient based class activation maps (GradCAM) \cite{selvaraju2017grad}, guided GradCAM \cite{selvaraju2017grad}, and GradCAM++ \cite{chattopadhay2018grad}. 
\end{enumerate}
This work considers the problem of gradient-free XAI. Below, we describe the details of the prominent gradient-free XAI approaches. 
In this description, we use the following notation. The input is denoted as $x_p$, the perturbed input samples are denoted as $\{x_p^i\}_{i=1}^Q$, where $Q$ denotes the number of perturbations,  $y_p^i$ denotes the output of the black-box model for the perturbed samples, $T$ denotes the target class, and $M$ denotes the multiplier mask. 

\subsubsection{Occlusion}
Occlusion methods originated from the   work on understanding neural networks~\cite{zeiler2014visualizing}. It is a sliding window based approach that hides   portions of the input image and studies the change in class activations to find the explanations. 
\subsubsection{RISE}
Randomized Input Sampling for Explanation (RISE)~\cite{petsiuk2018rise} is another perturbation based approach, where the perturbation samples $x_p^i$ are generated by masking out the input image using randomly generated binary masks. The masked image samples are passed through the black-box and the corresponding black-box scores for the target class $[y_p^i]_T$ are stored. The final explanation $E_{rise}(x_p,T)$ is computed as,
\begin{equation}
    E_{rise}(x_p,T) = \frac{\sum_{i=1}^{Q}x_p^i \dot [y_p^i]_T }{\sum_{i=1}^{Q}[y_p^i]_T}
\vspace{-0.01in}
\end{equation}
\subsubsection{SHAP}
SHapley Additive exPlanations~\cite{lundberg2017unified} is a game theoretic approach to explainability. It computes shapely values as explanation and proves the uniqueness of the solution. Their formulation assumes the explanation itself as a model, which is called as  the explanation model.
\subsubsection{LIME}
Local Interpretable Model-agnostic Explanations (LIME)~\cite{ribeiro2016should} is based on the   popular approximation in the  perturbation based methods - the local linear approximation. The local linear approximation provides explanation by exploiting two facts - a) the linear regression model is inherently interpretable, and b) the non-linear function can be approximated as a linear one within the local neighbourhood. 
The working details of the LIME model are shown in Figure~\ref{fig:lime-working-principle}.  
LIME first perturbs the local input to generate numerous neighbourhood samples, computes the corresponding black-box responses using the query access to black-box, and  fits a linear regression model using the mask and responses as inputs and targets,  respectively. 

\begin{figure}[t!]
    \centering
    \includegraphics[width=8cm, height=4.3cm]{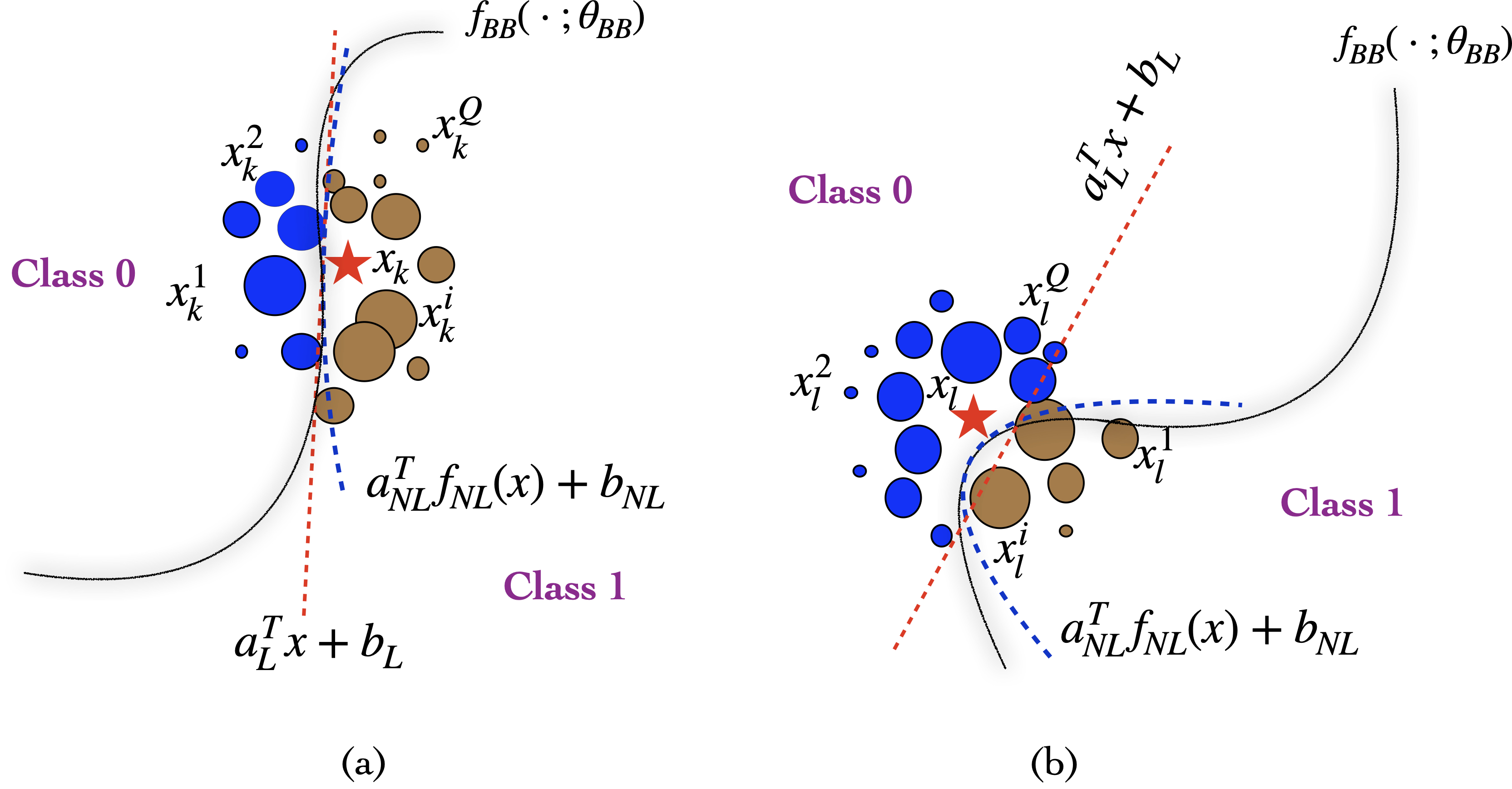}
    \caption{
    Let $f_{BB}(\cdot; \theta_{BB})$ be the trained black box neural network whose decision boundary (separating two classes) is shown by black solid lines in the figure. The decision boundary has different local curvatures near local inputs $x_k$ (figure (a)) and $x_l$ (figure (b)). (a) Locally linear assumption (red dashed line) is a good approximation of the black box at $x_k$. (b) As the decision boundary is less smooth near $x_l$, a linear assumption (red dashed line) is a bad local approximation of the black-box. A mild non-linearity (blue dashed line) can significantly reduce the approximation error.
    }
    \vspace{-0.1in}
    \label{fig:lime-non-linerity}
\end{figure}

The LIME approach has three major drawbacks, which we attempt to address in the proposed DAX framework.
\textbf{Limitations of linear approximation}: If the locality of black-box function near an input is not smooth, as shown in Figure~\ref{fig:lime-non-linerity}, the explanations are generated from an invalid   approximation of the black-box model.
\revision{A potential solution to overcome this scenario is to allow the local approximation to be non-linear. In the proposed DAX setting, we have designed a student network to approximate the local behavior. As shown in Figure~\ref{fig:lime-non-linerity}(b), even a mildly non-linear approximation can mitigate the approximation error to a large extent.}
\\\textbf{Input data manifold}: Images are associated with very high dimensionality ($\mathcal{O}$($10^4$) for $224$ X $224$ image). Local linear approximations are cumbersome to model on such high dimensional sub-spaces. As a way around, LIME segments the input image, and uses the segment locations as the input variables without working on the pixels. This leads to a drastic reduction in the input dimensionality  of $\mathcal{O}$($10^1$). The LIME \cite{ribeiro2016should} fits a linear model using binary information about masking (if a segment is masked-off or not), as shown in Figure~\ref{fig:lime-working-principle}.
This may bring  the reliability of the explanations into question. 
Narodytska et al. \cite{narodytska2019assessing} argues that the LIME generated explanations may not be accurate as it operates on a reduced subspace of the local input.
\revision{In the proposed DAX framework, we attempt to overcome this issue by directly operating on the raw input image space.}\\
\textbf{Imprecise explanations}: The explanation importance weights are obtained at the segment level. This may lead to imprecise explanations, as shown in Figure~\ref{fig:lime-working-principle}(d).
\revision{In the proposed DAX framework, the explanations are obtained at the pixel level, making them very precise. We also use the segment-anything-model (SAM) \cite{kirillov2023segment} to generate precise segments that can be used for the XAI setting.}


\section{Methodology}
\label{sec:methodology}
\subsection{Problem formulation}
We formalize the problem in a supervised classification setting. Let, $f_{BB}(\cdot; \theta_{BB})$ denote a trained deep neural network based classifier (black-box), which is  to be explained. The training dataset is denoted as $\{x_i\in X\sim D_X,\, z_i\in Z\sim D_Z\}_{i=1}^{N}$, with $x_i\in \mathcal{R}^D$ as inputs, and $z_i\in \mathbb{Z}$ denoting the one-hot labels. The pre-trained model parameters are denoted as $\theta_{BB}$, with $C$ classes. Thus, the model $f_{BB}(\cdot, \theta_{BB})$ has $C$ output nodes with softmax non-linearity,
\begin{equation}
    |y| = |f_{BB}(x; \theta_{BB})| = C;~ \, \sum_{i=1}^{C}[f_{BB}(x; \theta_{BB})]_i = 1, \, \forall x
\end{equation}
The problem of explaining the black-box is defined for a given input $x_p$.
For the target-class $T$, the goal is to explain the prediction $[y_p]_T$, given $x_p$, where $y_p = f_{BB}(x_p; \theta_{BB})$.
\subsection{Functional view of ``explanation"}
We pose explanation for the black-box  as a function,
\begin{equation}
E [{x_p}, T] = E(x_p, y_p |f_{BB}(\cdot; \theta_{BB}), T) \in \mathcal{R}^D
\label{eqn:func_view}
\end{equation}
where $E[{x_p}, T]$ denotes an explanation computed on $x_p$, in the form of weights assigned to each dimension of $x_p$ for the prediction of the target-class $T$, denoted as $[y_p]_T$. In particular, $E[{x_p},T]\in \mathcal{R}^D$ denotes a saliency  map (also referred to as explanation mask),  where $[E[{x_p}, T]]_k$, denotes the importance of $k$-th dimension of $x_p$ in generating the prediction,  $[y_p]_T$. 



\subsubsection{Neighbourhood sampling}
To generate the explanation for a local input, $x_p$, our approach  relies on neighbourhood samples at close proximity of $x_p$, obtained by perturbing $x_p$. As pointed out by Fong et. al.~\cite{fong2017interpretable}, the perturbations must be ``meaningful'' (having contextual meaning). For images, segmentation~\cite{vedaldi2008quick} is an algorithm that captures contextual meaning. Hence, we perturb image segments to obtain neighbourhood samples - similar to the perturbation strategy used by LIME~\cite{ribeiro2016should}. 
Let, the local input $x_p$ contain $S$ segments $\{u_1, u_2, ..., u_S\}$ obtained using an image segmentation algorithm. The segments are non-intersecting and exhaustive.
\begin{equation}
    u_i\in x_p \, \forall i; \, u_i\cap u_j = \phi, \, i\neq j; \, \bigcup_{i=1}^{S}u_i = x_p
\end{equation}
For generating $Q$  neighbourhood samples,   $x_p$ is randomly perturbed $Q$ times by randomly masking off a subset of the segments $\{u_i\}_{i=1}^S$. Let the neighbourhood samples be denoted as  $\{x_p^i\}_{i=1}^Q$, with the $i$-th   sample $x_p^i$ containing only a subset of the segments. Let the mask of indices be denoted by $I_i\in \mathbb{Z}^{SX1}$, where $[I_i]_j = \mathbbm{1}_{\{j\text{-th segment is masked-off}\}}$. 
Thus, the $i$-th neighbourhood sample of $x_p$ is given by,
\begin{equation}
    x_p^i = \bigcup_{\substack{j=1 \\ [I_i]_j = 0}}^{S}u_j 
\end{equation}
It is noteworthy that existing local approximation methods~\cite{ribeiro2016should} use the low dimensional $I_i$ as inputs to the local approximation model (Figure~\ref{fig:lime-working-principle}). 
In contrast, our approach directly operates on data samples, $x_p^i$, to generate the explanation.
\subsubsection{Explanation strategy}
The problem is to find the explanation $E[{x_p},T]$ given $Q$ neighbourhood samples $\{x_p^i\}_{i=1}^Q$,   query access to $f_{BB}(\cdot; \theta_{BB})$, and the target-class $T$. \\ 
\textbf{The learnable distillation approach} : 
Our approach has two learnable components - a \textit{mask generation network} and a \textit{student network}. The job of the mask generation network is to find the salient regions in the image (the explanation), while the student network tries to locally distill the black-box. 
\begin{figure*}[t!]
\centering
\includegraphics[width=18cm, height=8cm]{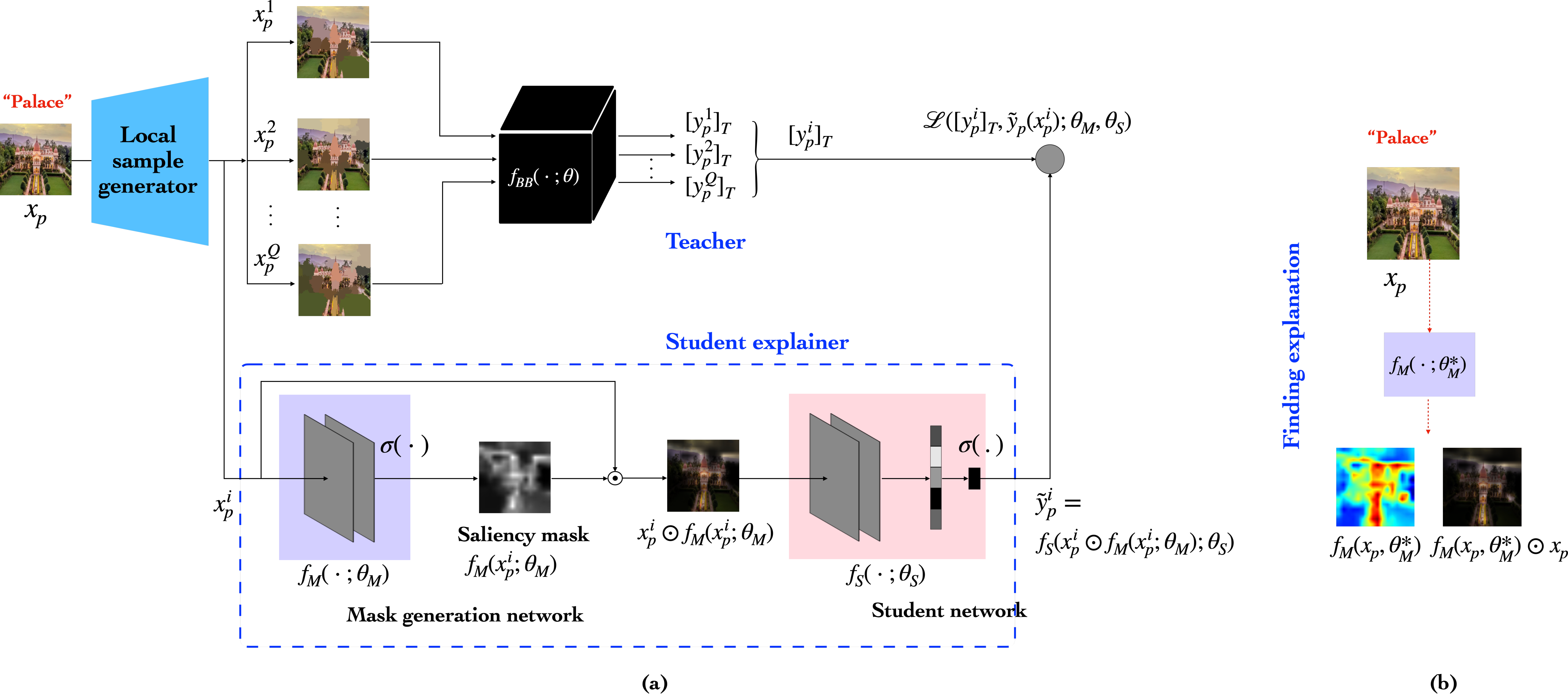}
\vspace{-0.1in}
\caption{\revision{The DAX framework. In part (a), we show the two components of the model, which are highlighted in the bottom row - i) Mask generation network, and ii) Student network. In part (b), the inference step is illustrated.} 
}
\vspace{-0.16in}
\label{fig:dame-architecture}
\end{figure*}
\subsection{DAX framework}
\subsubsection{``Explanation'' as the optimal multiplier}
\label{subsec:explanation-multiplier}
We look at the notion of explanation as an optimal multiplier of the input that causes minimum drop in the black-box prediction of the target-class. The true explanation, $E^* [x_p,T]$, assigns  higher weights to the dimensions of $x_p$ that are most important for the model, $f_{BB}$, to make the target-class prediction $[y_p]_T$. If $M_p^T\in \mathcal{R}^D$ is any input multiplier and $0\leq [M_p^T]_k\leq 1 \, \forall k$, then we obtain the optimal multiplier, $M_p^{T*}$, with the optimization,
\begin{equation}
    E^* [x_p,T] = M_p^{T*} =  \argmin_{M} \left([y_p]_T - f_{BB}(M\odot x_p; \theta_{BB})\right)^2
    \label{eqn:optimal-multiplier}
\end{equation}
Equation~(\ref{eqn:optimal-multiplier}) poses the problem of finding the explanation as a learnable approach, where explanations can be found by minimizing the mean squared error (MSE) between black-box output for the original input and the black-box  response for the ``perfectly'' masked input.
\subsubsection{Distillation to avoid gradient flow restriction}
Using the perturbation samples $\{x_p^i\}_{i=1}^{Q}$ and corresponding black-box responses $\{[y_p^i]_T\}_{i=1}^{Q} = \{[f_{BB}(x_p^i; \theta_{BB})]_T\}_{i=1}^{Q}$, one can optimize the explanation motivated by Equation~(\ref{eqn:optimal-multiplier}). 

However, \revision{in a gradient-free setting, gradient flow through black-box model is not available}. In order to perform the optimization described above without gradient-access, we propose a student distillation model, which locally approximates the black-box model. In this way, the distilled network can be used for learning the explanation, as the gradients can be computed. 

Thus, we have two optimization problems at hand - i) Finding a local approximation of the black-box using distillation and, ii) Finding a multiplier based explanation (Equation~\ref{eqn:optimal-multiplier}). We perform a joint optimization of these two sub-problems, as described in the following sub-sections. 
\subsubsection{Local distillation of black-box}
The black-box is locally distilled by approximating the black-box behavior at the vicinity of local input, $x_p$. The neighbourhood samples $\{x_p^i\}_{i=1}^{Q}$ and black-box responses $\{[y_p^i]_T\}_{i=1}^{Q}$ are used to train a small learnable network, $f_S(\cdot; \theta_S)$, where $ \theta_S$ denotes the model parameters of the student network. We minimize the MSE loss to learn the parameters of the student network, $f_{S}(x_p^i; \theta_{S})$, i.e.,
\begin{equation}
    \theta_S^* = \argmin_{\theta _S} \sum _{i=1}^Q \left([y_p^i]_T - [f_{S}(x_p^i; \theta_{S})]_T\right)^2
    \label{eqn:mse-loss-distillation}
\end{equation}

\subsubsection{Learnable explanation}
Interestingly, as seen in Equations (\ref{eqn:optimal-multiplier}) and (\ref{eqn:mse-loss-distillation}),   finding the explanation and local distillation are posed as the  minimization of objective functions based on the MSE loss. We propose to combine the two optimization problems with 
two   learnable networks, a mask generation network, $f_M(\cdot; \theta_M)$ and a student network, $f_S(\cdot; \theta_S)$. Now, minimization of the following loss
\begin{dmath}
    \mathcal{L}_{\mathbb {MSE}}(x_p^i, T ; \theta_M, \theta_S) =   
    \left[[y_p^i]_T - f_{S}(x_p^i\odot f_M \left(x_p^i; \theta_{M}) \right);\theta_S)\right]^2
    \label{eqn:mse-loss-learnable}
\end{dmath}
allows the estimation of the parameters $\{\theta_M, \theta_S\}$.
Here, $f_M(x_p; \theta_{M})$ is the multiplier mask generated by the mask generation network, while \revision{the product, $\boldsymbol{x_p} \odot \boldsymbol{f_M}(\boldsymbol{x_p}; \boldsymbol {\theta_{M}} )$, is the salient explanation output of the proposed DAX framework. } Let $\Tilde{y}_p(x_p^i) = f_{S}(x_p^i\odot f_M(x_p^i; \theta_{M});\theta_S)$ denote the output of the DAX model for the perturbed input $x_p^i$. 

Note that, optimizing based on Equation~(\ref{eqn:mse-loss-learnable}) can be ill-posed,  as it can lead to the trivial identity solution $f_M(x_p; \theta_{M})=\mathbb{I}$.  It can be avoided by using a regularized loss.  We propose an $\mathbb {L}_1$ loss, as discussed in Equation~(\ref{eqn:final-loss}) given below.
\begin{eqnarray}
    \{\theta_M^*, \theta_S^*\} = \argmin_{\theta_M, \theta_S}\mathcal{L}_{\mathbb {TOTAL}}(x_p, T ; \theta_M, \theta_S) \\   \revision{\mathcal{L}_{\mathbb {TOTAL}}(x_p, T ; \theta_M, \theta_S) = \sum_{i=1}^{Q} \gamma_i~ \mathcal{L}_{\mathbb {MSE}}(x_p^i, T ; \theta_M, \theta_S)} \nonumber \\   
   \revision{+  \lambda_1 \mathcal{L}_{\mathbb {SP}}(x_p^i; \theta_M)}    \revision{+ \lambda_2~ \mathcal{L}_{\mathbb {KL}}(x_p^i; \theta_M, \theta_S)} 
    \label{eqn:final-loss} 
 \end{eqnarray}
 \revision{where,}
 \begin{eqnarray}
\revision{\mathcal{L}_{\mathbb {SP}}(x_p^i; \theta_M)} & \revision{=} & \revision{ \sum_{i=1}^{Q} ||f_M(x_p^i; \theta_{M})||_{\mathbb {L}_1}} \\ 
\revision{\mathcal{L}_{\mathbb {KL}}(x_p^i; \theta_M, \theta_S)} & \revision{=} & \revision{  {\mathcal KL} \left(\{\textbf{q}_{\{[y_p^i]_T\}_{i=1}^Q} ~||~ \textbf{q}_{\{{\Tilde{y}_p\left(x_p^i\right)\}_{i=1}^Q}} \right)} \nonumber 
\end{eqnarray}
The first term of the total loss is the MSE loss (defined in Equation (\ref{eqn:mse-loss-learnable})), the second term is the $\mathbb{L}1$ penalty loss on the mask explanations, and the third term is the $KL$-divergence between distribution ($\textbf{q}$) of black-box model scores ($\{[y_p^i]_T\}_{i=1}^Q$) and the distillation output scores ($\{{\Tilde{y}_p\left(x_p^i\right)\}_{i=1}^Q}$). 

The $KL$-divergence loss helps in bringing  the distributions of DAX scores closer to the black-box scores, which regularizes the MSE loss. The $\gamma_i$ weighs the ``closeness'' of the neighbourhood samples  to the actual input $x_p$, i.e., $\left(\gamma_i \propto \frac{1}{||x_p - x_p^i||_{L2}}\right)$. Further,  $\Tilde{y}_p\left(x_p^i\right)$ is defined in the description below Equation~(\ref{eqn:mse-loss-learnable}), and $\lambda_1$, $\lambda_2$ are hyper-parameters that are set based on a held-out validation set.

The distribution $\textbf{q}$ is computed at a batch-level using a histogram.
The optimization of the total loss is performed in iterative fashion using gradient descent. The neighbourhood samples are divided into training and validation set following standard machine learning protocols. In our experiments, the number of local neighbourhood samples and batch size used for DAX saliency learning are $6000$ and $64$, respectively. Also, the hyper-parameters $\lambda_1$ and $\lambda_2$ are selected as $0.001$ and $0.02$ respectively, based on experiments on the validation set.\\
\revision{We have also experimented with a quadratic counterfactual loss for encouraging sparsity, instead of the $\mathcal{L}_{\mathbb {SP}}(x_p^i; \theta_M)$. It minimizes the agreement between the student model scores and the black-box scores when the mask $f_M(x_p^i; \theta_{M});\theta_S)$ is complemented.}
\begin{equation}
    \revision{\mathcal{L}_{\mathbb {CNT}}(x_p^i, T; \theta_M, \theta_S)=\left([y_p^i]_T - f_{S}(x_p^i\odot f_M^C \left(x_p^i; \theta_{M}) \right);\theta_S)\right)^2} \nonumber 
\end{equation}
\revision{where,} 
\begin{equation}
    \revision{   f_M^C (x_p^i; \theta_{M}) = (\mathbf{1} - f_M \left(x_p^i; \theta_{M}) \right)), ~~\mathbf{1}_{ij} = 1, \forall i,j}
\end{equation}
\revision{In our experiments, $\mathcal{L}_{\mathbb {CNT}}$ is seen to be superior to $\mathcal{L}_{\mathbb {SP}}$, resulting in significantly better performance on all the metrics, even in absence of $\mathcal{L}_{\mathbb {KL}}$. Different variants of DAX loss function  that are used in experiments is listed in Table~\ref{tab:dax-variants}.}
\begin{figure}[t!]
\centering
\includegraphics[width=0.47\textwidth, height=0.25\textwidth]{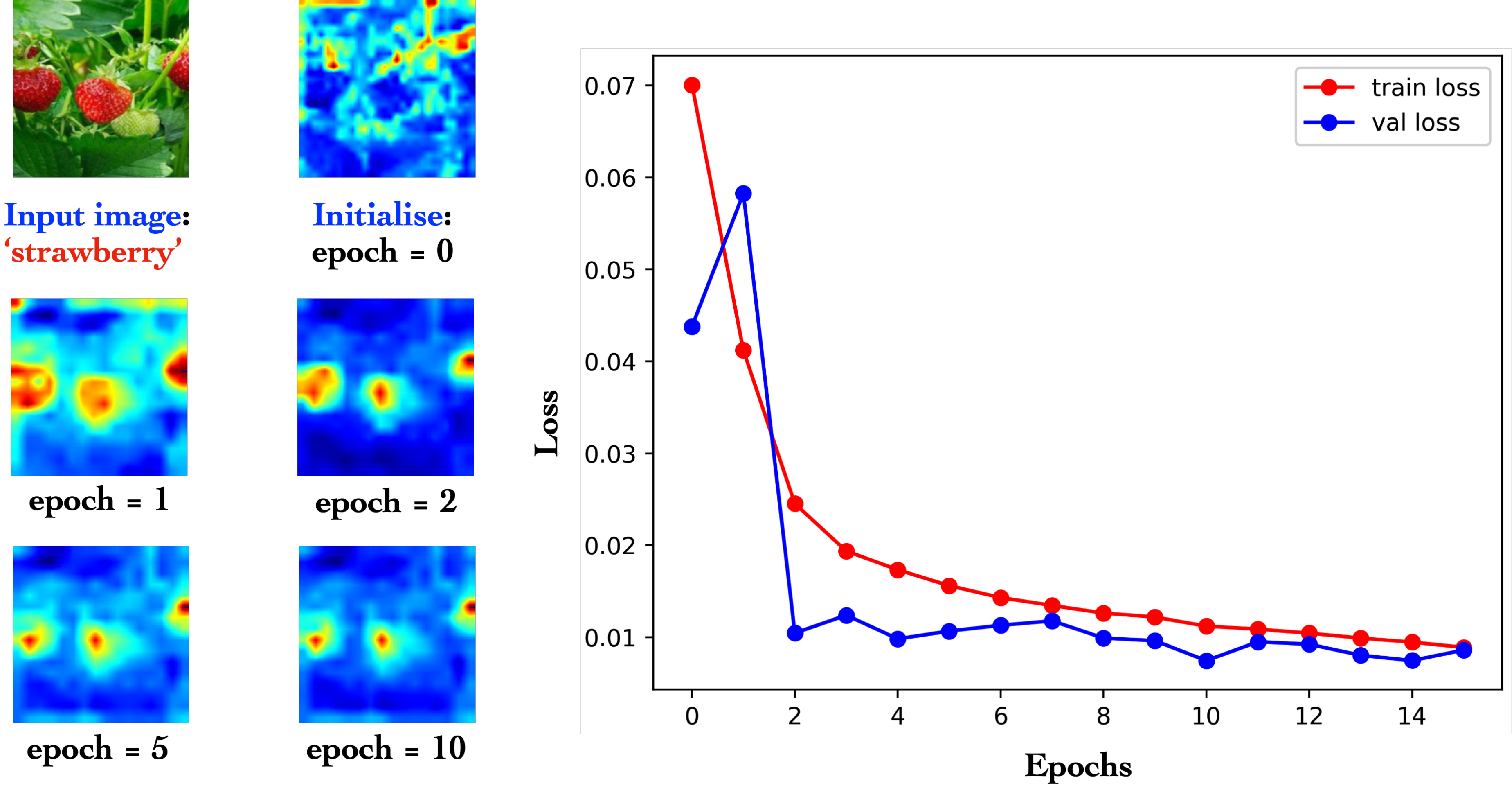}
\caption{The learnable explanation converges in less than $10$ epochs. The (left) panel shows the mask generated in different epochs, and (right) panel shows the training and validation loss curves. 
}
\vspace{-0.1in}
\label{fig:epochwise-expl-and-loss}
\end{figure}
\subsection{Model architecture}
We use a two layer convolutional network (CNN) with sigmoid activation for mask generation network,  $f_M(\cdot; \theta_{M})$. The student network, $f_S(\cdot; \theta_{S})$ consists of $2$ CNN layers followed by a fully-connected layer with sigmoid activation, as shown in Figure~\ref{fig:dame-architecture}. 

The sigmoid function at the output of $M = f_M(\cdot;\theta _M)$ ensures that the mask multiplier satisfies $M_{ij}\in[0,1]$,  while the sigmoid activation at the output of the student network ensures that the distilled scores ($\Tilde{y}_p$) and the target scores $y_p$ are in the same numerical range.

After the training is complete, the explanation can be obtained as, 
\begin{equation}
    E[{x_p},T] = x_p \odot f_M(x_p ; \theta_{M}) 
\end{equation}
and the student network $f_S(\cdot; \theta_{S}^*)$ can be discarded. The steps to find the explanation\footnote{The implementation of the DAX model is available in \url{https://github.com/iiscleap/DAX}} using proposed approach are outlined in Algorithm~\ref{algo:xai-algo}.
\RestyleAlgo{ruled}
\begin{algorithm}
\caption{Finding explanation}\label{alg:two}
\textbf{Inputs:} $f_{BB}(\cdot; \theta_{BB})$, $x_p$, $T$\\
\textbf{Output:} $E[{x_p}, T]$\\
\textbf{Training:}
Initialization:\\ \{$n, Q\} \gets \{no.\, of\, epochs, no.\, of\, perturbations$\};\\
$\{\theta_H, \theta_D\} \gets random$;\\
$epoch \gets 0$\\
Obtain $\{x_p^i\}_{i=1}^{Q}$  and $\{[y_p^i]_T\}_{i=1}^Q=\{[f_{BB}(x_p^i; \theta_{BB})]_T\}_{i=1}^Q$\\
\While{$epoch~ k \leq n$}{
   Compute $f_M(x_p^i;\, \theta_M)$ ;\\
  Compute $\Tilde{y}_p(x_p^i) = f_{S}(x_p^i\odot f_M(x_p^i; \theta_{M});\theta_S)$ ;\\
  Compute $\mathcal{L}_{\mathbb {TOTAL}}$ using Equation (\ref{eqn:final-loss});
  $\left(\theta _M, \theta _S \right)^{(k)} \gets \left (\theta _M, \theta _S \right )^{(k-1) } - \eta \nabla \mathcal{L}^{(k-1)}_{\mathbb {TOTAL}}$;\\
  $k = k + 1$;
}
\textbf{Inference:}
$E [{x_p}, T] = x_p\odot f_M(x_p; \theta_{M})$ 
\label{algo:xai-algo}
\end{algorithm}

\revision{We also explore the application of segment-anything-model (SAM) ~\cite{kirillov2023segment} as the segmentation algorithm for the proposed DAX framework.
 The SAM framework consists of an image encoder, an optional prompt encoder and a mask decoder. We have used the SAM setup without any prompts and used the segmentation maps to generate perturbations.
 The different variants of the proposed DAX framework are mentioned in Table~\ref{tab:dax-variants}. 
 }
The epoch-wise progression of mask generation and the corresponding loss curves are shown in Figure~\ref{fig:epochwise-expl-and-loss}, for the example image of strawberry. The saliency explanation converges  quickly, within $10$ epochs. The epoch-wise loss values on training and validation sets are also shown.
\begin{table}[t!]
\centering
\bgroup
\def\arraystretch{1.2}%
\begin{tabular}{ccclr}
\toprule[0.5pt]
\multicolumn{1}{c}{{\color[HTML]{000000} }}                                                                                   & {\color[HTML]{000000} \textbf{}}            & \multicolumn{3}{c}{\textbf{\revision{Components}}}                                                                                                                                                          \\ \cline{3-5} 
\multicolumn{1}{c}{\multirow{-2}{*}{{\color[HTML]{000000} \textbf{\begin{tabular}[c]{@{}c@{}}\revision{DAX} \\ \revision{variants}\end{tabular}}}}} &                                             & \textbf{\begin{tabular}[c]{@{}c@{}}\revision{Segmentation algo. } \\  \end{tabular}} & \textbf{} & \multicolumn{1}{c}{\textbf{\begin{tabular}[c]{@{}c@{}}\revision{Loss}\\ \end{tabular}}} \\ \cline{1-1} \cline{3-3} \cline{5-5} 
{\color[HTML]{000000}  \revision{V1}}                                                                                                     & \multicolumn{1}{c}{{\color[HTML]{000000} }} & \revision{Quick shift \cite{vedaldi2008quick}}                                                                                    &           & \revision{$\mathcal{L}_{MSE}+\lambda_1 \mathcal{L}_{L_1}+\lambda_2 \mathcal{L}_{KL}$}                                                                                      \\
{\color[HTML]{000000} \revision{V2}}                                                                                                     & \multicolumn{1}{c}{{\color[HTML]{000000} }} & \revision{Quick shift \cite{vedaldi2008quick} }                                                                                    &           & \revision{$\mathcal{L}_{MSE}+\lambda _3  \mathcal{L}_{CNT}$}                                                                                      \\
\revision{V3}                                                                                                                            &                                             & \revision{SAM \cite{kirillov2023segment}}                                                                                          &           & \revision{$\mathcal{L}_{MSE}+\lambda _4  \mathcal{L}_{CNT}$}                                                                                      \\ 
\bottomrule[0.5pt]
\end{tabular}
\egroup
\vspace{0.04in}
\caption{\revision{Different variants of DAX formulation studied.}}
\label{tab:dax-variants}
\vspace{-0.3in}
\end{table}
\begin{figure*}[t!]
    \centering
    \includegraphics[width=18cm, height=17.5cm]{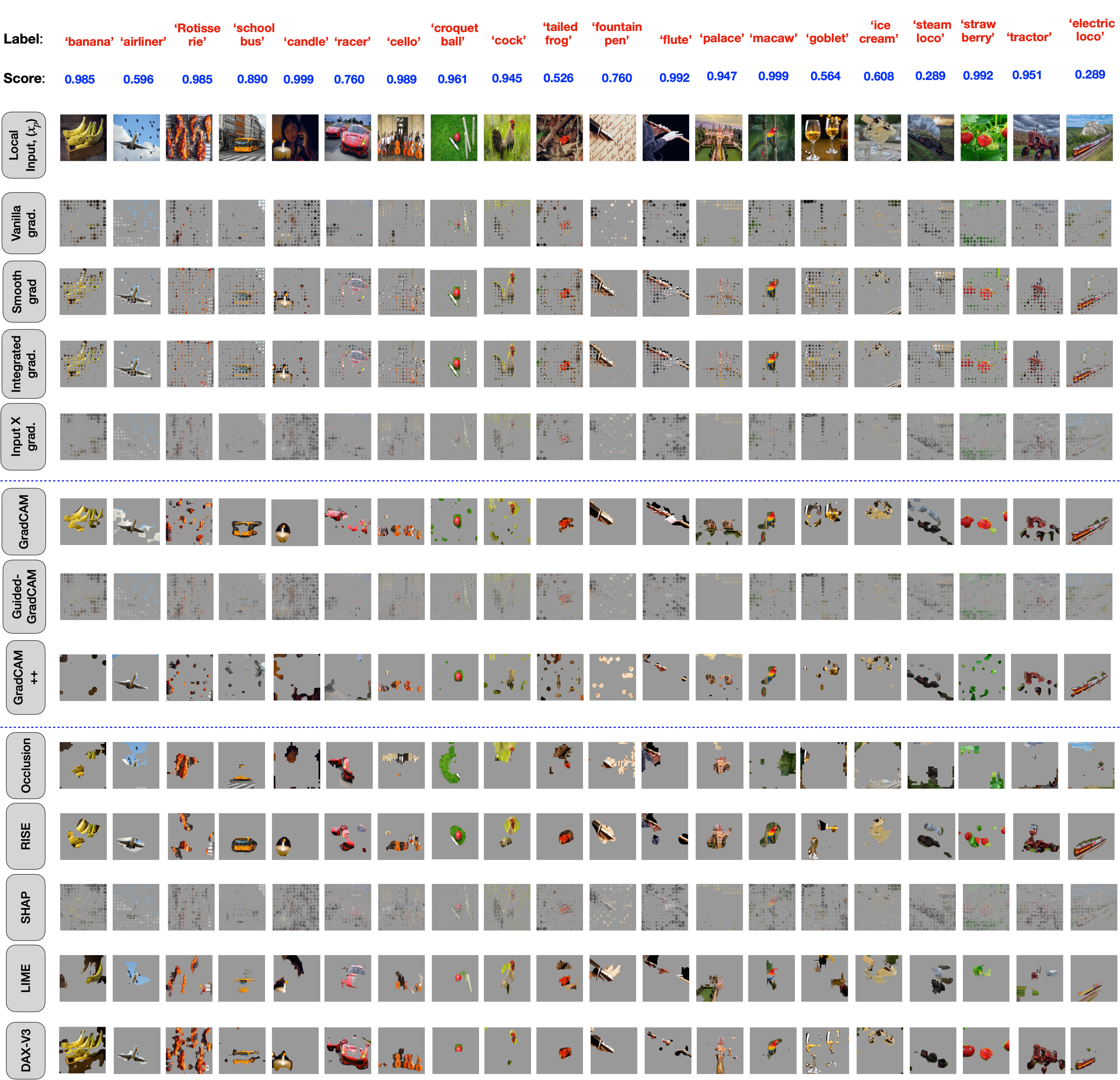}
    \vspace{-0.1in} 
    \caption{Comparison of explanations generated by DAX and $11$ other existing approaches for the ViT-b16 black-box model on the ImageNet dataset.}
    \label{fig:vision-results-all}
    \vspace{-0.15in}
    
\end{figure*}

\begin{table*}[h!]
\centering
\resizebox{0.9\linewidth}{!}{
\bgroup
\def\arraystretch{1.2}%
\begin{tabular}{lcccccccc}
\toprule[0.9pt]
\multicolumn{3}{c}{{\color[HTML]{000000} }}                                                                                 & \multicolumn{1}{l}{{\color[HTML]{000000} }} & \multicolumn{5}{c}{{\color[HTML]{000000} \textbf{Black box network}}}                                                                                                                                                                                                                                                                          \\ \cline{5-9} 
\multicolumn{3}{c}{\multirow{-2}{*}{{\color[HTML]{000000} \textbf{XAI Methods}}}}                                         & \multicolumn{1}{l}{{\color[HTML]{000000} }} & \multicolumn{2}{c}{{\color[HTML]{000000} ResNet 101}}                                                                                      & \multicolumn{1}{l}{{\color[HTML]{000000} }} & \multicolumn{2}{c}{{\color[HTML]{000000} ViT-base-16}}                                                                                     \\ \cline{1-3} \cline{5-6} \cline{8-9} 
{\color[HTML]{000000} Name}                  & {\color[HTML]{000000} \begin{tabular}[c]{@{}c@{}}Gradient\\ free  \end{tabular}} & {\color[HTML]{000000} \begin{tabular}[c]{@{}c@{}}Model\\ agnostic  \end{tabular}} & \multicolumn{1}{l}{{\color[HTML]{000000} }} & {\color[HTML]{000000} \begin{tabular}[c]{@{}c@{}}Mean \\ IoU (\%) ($\uparrow$)  \end{tabular}}        & {\color[HTML]{000000} \begin{tabular}[c]{@{}c@{}}Mean IoU (\%) on\\ incorrect label ($\downarrow$)\end{tabular}} & \multicolumn{1}{l}{{\color[HTML]{000000} }} & {\color[HTML]{000000} \begin{tabular}[c]{@{}c@{}}Mean \\ IoU (\%) ($\uparrow$)  \end{tabular}}       & {\color[HTML]{000000} \begin{tabular}[c]{@{}c@{}}Mean IoU (\%) on\\ incorrect label ($\downarrow$)\end{tabular}} \\ \cline{1-3} \cline{5-9} 
{\color[HTML]{000000} Vanilla gradient~\cite{simonyan2014visualising}}      & {\color[HTML]{000000} $\times$}            & {\color[HTML]{000000} $\checkmark$}            & {\color[HTML]{000000} }                     & {\color[HTML]{000000} 24.8 (8.1)}           & {\color[HTML]{000000} 0.5e-1}                                                                & {\color[HTML]{000000} }                     & {\color[HTML]{000000} 19.6 (5.3)}           & {\color[HTML]{000000} 0.6e-1}                                                                \\
{\color[HTML]{000000} Smooth-grad~\cite{smilkov2017smoothgrad}}           & {\color[HTML]{000000} $\times$}            & {\color[HTML]{000000} $\checkmark$}            & {\color[HTML]{000000} }                     & {\color[HTML]{000000} 34.5 (11.7)}          & {\color[HTML]{000000} 1.1e-1}                                                                & {\color[HTML]{000000} }                     & {\color[HTML]{000000} 20.6 (6.4)}           & {\color[HTML]{000000} 0.8e-1}                                                                \\
{\color[HTML]{000000} Integrated gradients~\cite{sundararajan2017axiomatic}}  & {\color[HTML]{000000} $\times$}            & {\color[HTML]{000000} $\checkmark$}            & {\color[HTML]{000000} }                     & {\color[HTML]{000000} 20.1 (8.5)}           & {\color[HTML]{000000} 1.1e-1}                                                                & {\color[HTML]{000000} }                     & {\color[HTML]{000000} 19.3 (15.4)}          & {\color[HTML]{000000} 1.3e-1}                                                                \\
{\color[HTML]{000000} Input x gradient~\cite{shrikumar2017learning}}      & {\color[HTML]{000000} $\times$}            & {\color[HTML]{000000} $\checkmark$}            & {\color[HTML]{000000} }                     & {\color[HTML]{000000} 19.9 (8.7)}           & {\color[HTML]{000000} 0.3e-1}                                                                & {\color[HTML]{000000} }                     & {\color[HTML]{000000} 19.2 (15.4)}          & {\color[HTML]{000000} 0.5e-1}                                                                \\ \hline
{\color[HTML]{000000} GradCAM~\cite{selvaraju2017grad}}               & {\color[HTML]{000000} $\times$}            & {\color[HTML]{000000} $\times$}             & {\color[HTML]{000000} }                     & {\color[HTML]{000000} 39.6 (13.9)}          & {\color[HTML]{000000} 2.8e-1}                                                                & {\color[HTML]{000000} }                     & {\color[HTML]{000000} 17.1 (11.9)}          & {\color[HTML]{000000} 2.6e-1}                                                                \\
{\color[HTML]{000000} Guided GradCAM~\cite{selvaraju2017grad}}        & {\color[HTML]{000000} $\times$}            & {\color[HTML]{000000} $\times$}             & {\color[HTML]{000000} }                     & {\color[HTML]{000000} 19.0 (10.1)}          & {\color[HTML]{000000} 0.8e-1}                                                                & {\color[HTML]{000000} }                     & {\color[HTML]{000000} 19.2 (15.4)}          & {\color[HTML]{000000} 0.5e-1}                                                                \\
{\color[HTML]{000000} GradCAM++~\cite{chattopadhay2018grad}}             & {\color[HTML]{000000} $\times$}            & {\color[HTML]{000000} $\times$}             & {\color[HTML]{000000} }                     & {\color[HTML]{000000} 39.7 (14.0)}          & {\color[HTML]{000000} 2.9e-1}                                                                & {\color[HTML]{000000} }                     & {\color[HTML]{000000} 10.5 (7.7)}           & {\color[HTML]{000000} 2.6e-1}                                                                \\ \hline
{\color[HTML]{000000} Occlusion~\cite{zeiler2014visualizing}}             & {\color[HTML]{000000} $\checkmark$}           & {\color[HTML]{000000} $\checkmark$}            & {\color[HTML]{000000} }                     & {\color[HTML]{000000} 29.6  (11.7)}         & {\color[HTML]{000000} 3.5e-1}                                                                & {\color[HTML]{000000} }                     & {\color[HTML]{000000} 28.5 (12.1)}          & {\color[HTML]{000000} 3.4e-1}                                                                \\
{\color[HTML]{000000} RISE~\cite{petsiuk2018rise}}                  & {\color[HTML]{000000} $\checkmark$}           & {\color[HTML]{000000} $\checkmark$}            & {\color[HTML]{000000} }                     & {\color[HTML]{000000} 33.7 (11.5)} & {\color[HTML]{000000} 4.5e-1}                                                                & {\color[HTML]{000000} }                     & {\color[HTML]{000000} {31.1 (11.8)}}    & {\color[HTML]{000000} 4.1e-1}                                                                \\
{\color[HTML]{000000} SHAP~\cite{lundberg2017unified}}                  & {\color[HTML]{000000} $\checkmark$}           & {\color[HTML]{000000} $\checkmark$}            & {\color[HTML]{000000} }                     & {\color[HTML]{000000} 22.4 (9.3)}           & {\color[HTML]{000000} 1.5e-1}                                                                & {\color[HTML]{000000} }                     & {\color[HTML]{000000} 19.3 (10.3)}          & {\color[HTML]{000000} 1.2e-1}                                                                \\
{\color[HTML]{000000} LIME~\cite{ribeiro2016should}}                  & {\color[HTML]{000000} $\checkmark$}           & {\color[HTML]{000000} $\checkmark$}            & {\color[HTML]{000000} }                     & {\color[HTML]{000000} 29.2 (10.9)}          & {\color[HTML]{000000} 3.2e-1}                                                                & {\color[HTML]{000000} }                     & {\color[HTML]{000000} 26.8 (11.4)}          & {\color[HTML]{000000} 4.0e-1}                                                                \\ \hdashline
{\color[HTML]{000000} \textbf{\revision{DAX-V1}}} & {\color[HTML]{000000} $\checkmark$}  & {\color[HTML]{000000} $\checkmark$}   & {\color[HTML]{000000} }                     & {\color[HTML]{000000} {33.3 (12.1)}}    & {\color[HTML]{000000} 2.8e-1}                                                                & {\color[HTML]{000000} }                     & {\color[HTML]{000000} 31.4 (11.9)} & {\color[HTML]{000000} 3.1e-1}                                                                \\
{\color[HTML]{000000} \textbf{\revision{DAX-V2}}} & {\color[HTML]{000000} $\checkmark$}  & {\color[HTML]{000000} $\checkmark$}   & {\color[HTML]{000000} }                     & {\color[HTML]{000000} { {\revision{34.7 (11.9)}}}}    & {\color[HTML]{000000} \revision{2.7e-1}}                                                                & {\color[HTML]{000000} }                     & {\color[HTML]{000000}  {\revision{32.5 (11.5)}}} & {\color[HTML]{000000} \revision{3.1e-1}}                                                                \\
{\color[HTML]{000000} \textbf{\revision{DAX-V3}}} & {\color[HTML]{000000} $\checkmark$}  & {\color[HTML]{000000} $\checkmark$}   & {\color[HTML]{000000} }                     & {\color[HTML]{000000} {\textbf{\revision{36.9 (11.6)}}}}    & {\color[HTML]{000000} \revision{2.5e-1}}                                                                & {\color[HTML]{000000} }                     & {\color[HTML]{000000} \textbf{\revision{34.1 (8.9)}}} & {\color[HTML]{000000} \revision{3.5e-1}}                                                                \\
\bottomrule[0.9pt]
\end{tabular}
\egroup
}
\vspace{0.1in}
\caption{The intersection-over-union (IoU) (\%) with associated standard deviation results for $11$ baseline approaches and the proposed DAX framework on $1246$ (and $1321$) images, which were correctly classified by the black-box ResNet-101 (and ViT-b16) model. The properties of the XAI methods are also highlighted, and the IoU for randomised incorrect labels are also given.
}
\label{tab:vision-results-iou}
\vspace{-0.2in}
\end{table*}

\section{Results}
\label{sec:results}
\subsection{Computer vision: object classification task}

\subsubsection{Qualitative analysis}
We present a diverse set of input images and the compute saliency explanations to enable extensive qualitative analysis. We compare the DAX explanations with those generated by $11$ other popular XAI approaches, belonging to $3$ different categories, to allow a strong benchmark comparison. For this analysis, the inputs that were correctly predicted by the black-box were considered.\\
\textbf{Task}: The object detection task is considered for this analysis. Given an input image, the task is to predict the class of the object present in the image.  Images are randomly selected from the evaluation set of ImageNet dataset~\cite{deng2009imagenet}.\\
\textbf{Black-box}: To analyze the explainability of the state-of-the-art models, a vision transformer~\cite{dosovitskiy2021an} of patch size $16$ (ViT-b$16$) is used as the black-box model, which was pre-trained on the ImageNet dataset.  
Figure~\ref{fig:vision-results-all} shows the explanations generated by our approach and  $11$ other popular XAI approaches. As seen here, the DAX model is able to generate consistent explanations for most of the classes shown here.

\subsubsection{Quantitative analysis}
\begin{figure}[t!]
\centering
\includegraphics[width=0.44\textwidth, height=0.2\textwidth]{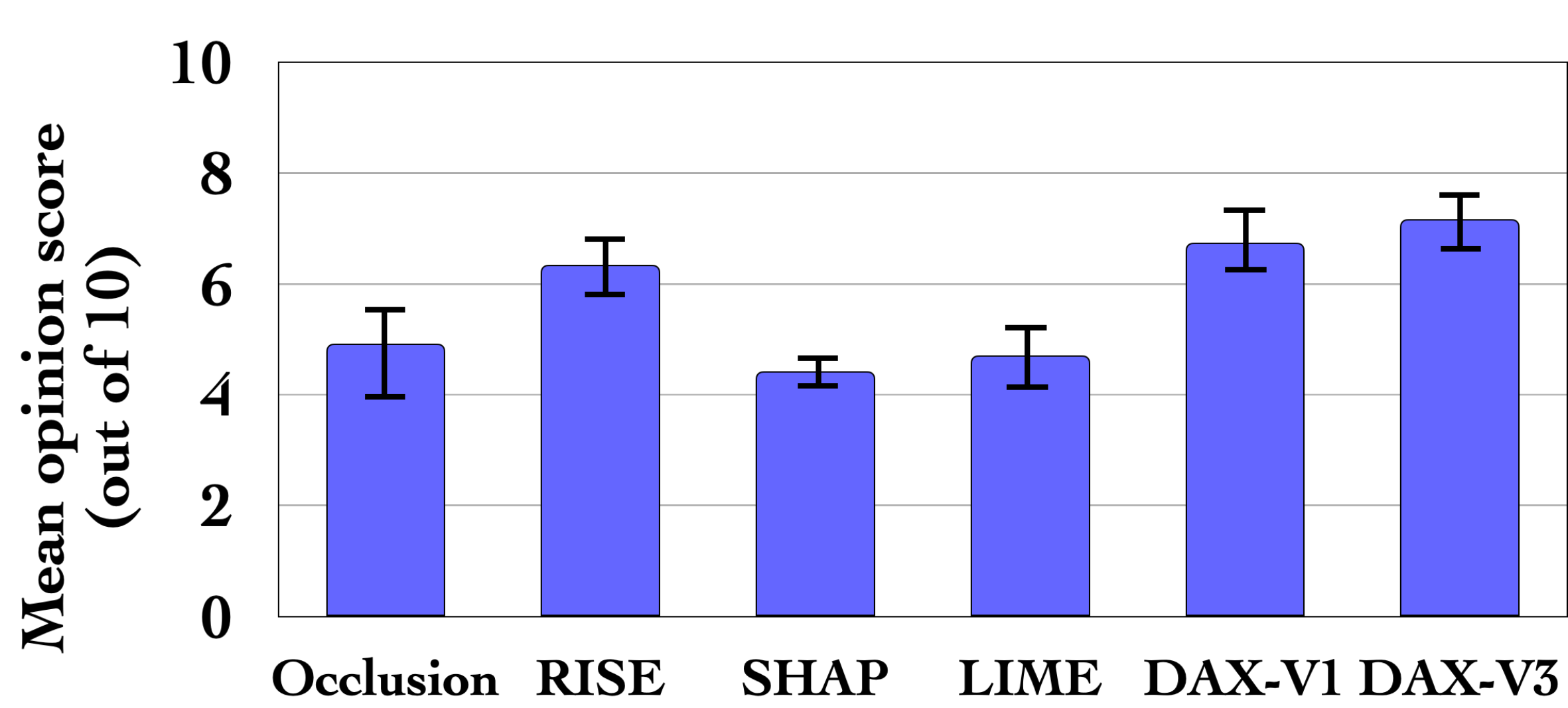}
\caption{\revision{Mean human opinion score (scale of $10$) for the explanation quality among different   XAI approaches collected from $30$ subjects, with each subject visualizing $25$ different image examples. DAX-V1 and DAX-V3 were preferred over others statistically significantly ($p<0.05$ in paired t-test).}
}
\vspace{-0.1in}
\label{fig:explanation-quality-study}
\end{figure}

\begin{figure}[t!]
\centering
\includegraphics[width=0.43\textwidth, height=0.37\textwidth]{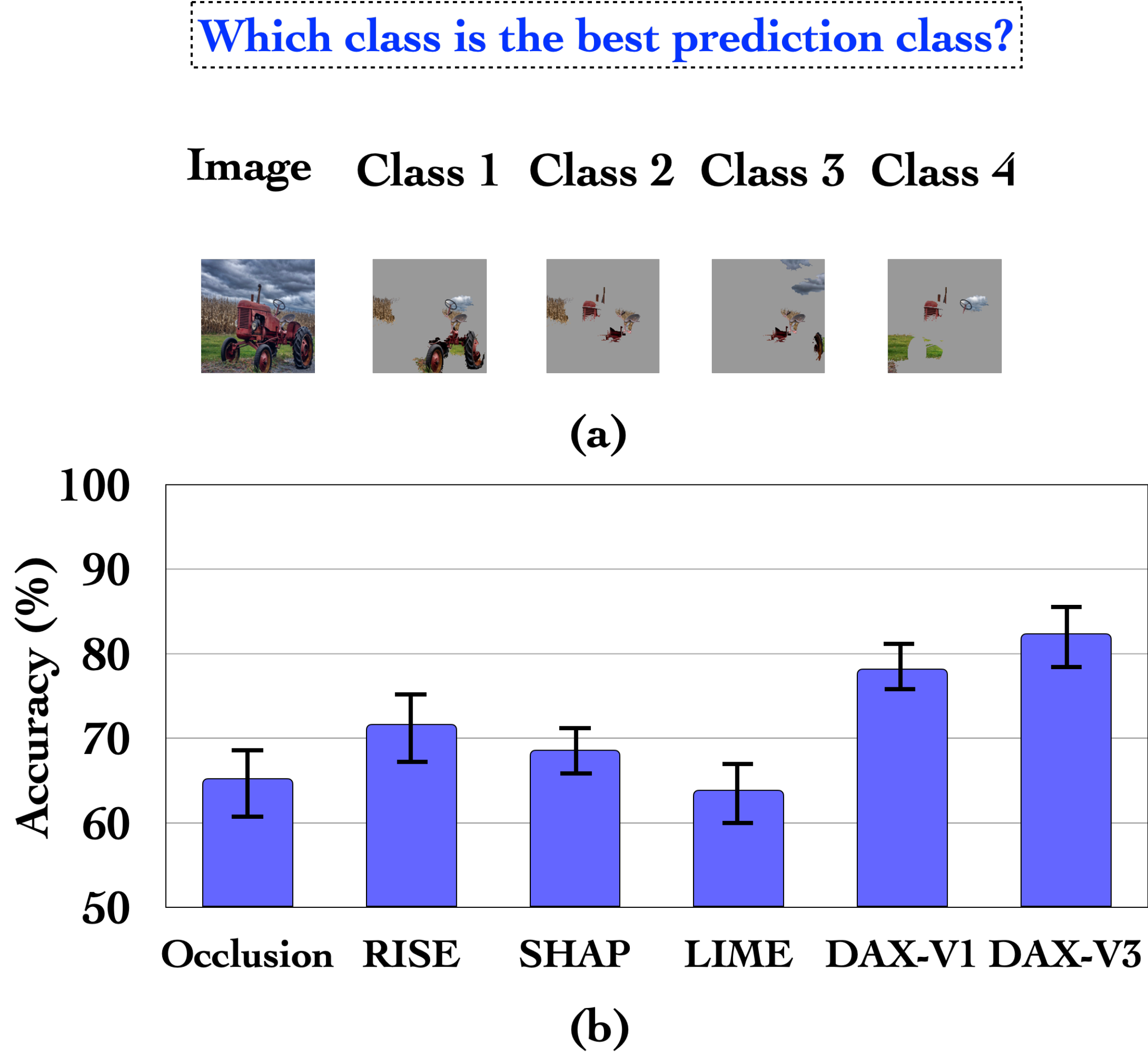}
\caption{\revision{The end-user utility study carried out for $6$ XAI methods according to HIVE~\cite{kim2022hive} guidelines. The study consists of $30$ human subjects. (a) An example of the questions provided to the subjects, and (b) Average accuracy of predicting the target class using the XAI map.}
}
\label{fig:end-user-study}
\end{figure}

Most of the prior explainability works only report the qualitative analysis on a selected subset of images. While this may provide visual insight, the statistical justification of the explanation quality is lacking. In this paper, we also present statistical performance of our approach on a large dataset. For extensive benchmark comparisons, we also evaluate $11$ other existing methods in the same setting.
\\\textbf{Task}: We use the Pascal visual object classification (VOC) dataset~\cite{everingham2015pascal}, comprising of images with $20$ object classes. It has two sub-parts - classification dataset (larger set having $5717$ images) and segmentation dataset (smaller set having $1464$ images). The images in the segmentation dataset also have the human annotated ground truth segmentation masks corresponding to the objects. We use the classification dataset for fine-tuning the models, while the segmentation dataset is used in our evaluations. \\
\textbf{Black-box}: We perform the statistical analysis separately on two different black-box models to analyze the performance consistency,  a) The ResNet-$101$~\cite{he2016deep} and, b) the ViT-b$16$~\cite{dosovitskiy2021an}. Both the black-box models are pre-trained on ImageNet~\cite{deng2009imagenet}. We fine-tuned the black-box models  on classification dataset of Pascal VOC, with $20$ classes. After fine-tuning, the ResNet-$101$ and ViT-b$16$ achieved an accuracy of $86.0\%$ and $91.2\%$, respectively on the segmentation set.
\\\textbf{Metric of explanation - IoU}: The   intersection-over-union (IoU) metric is used to quantify the explanation quality.
We use the ground truth object segmentation and compare this with the mask generated by the model ($M = f_M(\cdot ; \theta _M)$). For this IoU computation, the mask $M$ is converted to a binary level by thresholding it. A threshold of $\mu_M+\sigma_M$ is used for the purpose, where $\mu_M$ and $\sigma_M$ are the mean and standard deviation of the values in the matrix $M$, respectively. 
The IoU metric reflects how much the generated explanations  aligns with human annotated ground truth of the object classes. \\
\revision{
\\\textbf{Metric of explanation - Deletion AUC}:
we also computed the deletion AUC performance (proposed in RISE \cite{petsiuk2018rise}) as the metric. In this setting, the salient regions of the image, provided by the XAI method, are masked away and the masked image is used as input to the black-box model. The masking is performed progressively  with descending order of importance of regions provided by the XAI method. The final masked input corresponds to removing the entire saliency region identified by the XAI method. The drop in target class accuracy, for each level of masking, is measured. The receiver-operating-curve (ROC) is computed with the percentage of masking against the drop in target class accuracy. A lower value of deletion AUC is preferable. 
Table ~\ref{tab:deletion-auc-voc} illustrates that the deletion AUC is the best for the  the proposed method (DAX-V3), similar to the trends seen with the IoU metric.
}
\\\textbf{Explanation accuracy}: Table~\ref{tab:vision-results-iou} shows the statistical performance of our approach and $11$ other popular XAI approaches on the segmentation dataset of Pascal VOC. The analysis is performed on a total of $1246$ and $1321$ images for the ResNet-101 and ViT model respectively, belonging to the $20$ classes, which were  correctly classified by the black-box model. Our approach, that attempts to overcome limitations of LIME, is seen to achieve significant improvements ($4.1\%$ in terms of mean IoU accuracy over LIME). \\
\revision {
\textbf{Subjective evaluation}: we performed two sets of subjective listening tests to allow robust evaluation of the XAI models. 
\begin{enumerate}
    \item In the first evaluation, we performed a passive task, where human participants visualized  the original image with the ground-truth target class.  They were tasked to rate the different XAI saliency maps in terms of explaining the target object in the given image. The saliency maps of different XAI methods were presented in random order.  The mean opinion score (MoS) for the different XAI models   with $30$ human subjects is shown in Fig.~\ref{fig:explanation-quality-study}. 
    \item In the second evaluation, an active human study was conducted according to the guidelines mentioned in human interpretability of visual explanations (HIVE) \cite{kim2022hive}. In this setup, shown in Fig.~\ref{fig:end-user-study}, the original image with the saliency maps are shown and the raters are asked to predict the object class that best describes the saliency map generated by the chosen XAI method. The rating is compared with the ground truth class and the   prediction accuracy with respect to the target class is used as measure of the subjective quality of the XAI.
    \end{enumerate}
In   these studies, both the DAX frameworks are consistently seen to improve over the prior works (statistically significant improvements confirmed with a pairwise t-test, with $p<0.05$).  Further, DAX-V3 is seen to outperform the DAX-V1 setting.  
}

In terms of benchmark comparison, for ResNet model, many of the gradient-access based approaches generate good IoU values (Table~\ref{tab:vision-results-iou}). The highest IoU is obtained using GradCAM++, which is a specially designed method for CNN based models using gradient-access.  Among the model-agnostic gradient-free methods, RISE and DAX approach perform better than all the other approaches.

However, for ViT black-box, which is a attention based architecture, all the gradient-based methods elicit a significant degradation in IoU values.  The DAX approach performs the best among the $12$ methods compared here. It provides a $4.6\%$ IoU gain over the LIME approach.
\\\textbf{Explanation sensitivity}: The explanation for a black-box is sought for a target-class $T$. The explanations should significantly change if $T$ changes, a measure we call as \textit{sensitivity} of explanations. To quantify this, while seeking explanations, we randomly pick one of the incorrect classes. In such a scenario, a lower IoU value is preferred, as this would indicate a discriminative explanation.   In Table~\ref{tab:vision-results-iou}, the DAX framework is seen to provide improved sensitivity over the RISE approach.

\begin{table}[t!]
\centering
\resizebox{0.85\linewidth}{!}{
\bgroup
\def\arraystretch{1.28}
\begin{tabular}{llcllrl}
\hline
{\color[HTML]{000000} }                                       & {\color[HTML]{000000} \textbf{}}            & \multicolumn{5}{c}{\textbf{\revision{Deletion-AUC ($\downarrow$)}}}                                                               \\ \cline{3-7} 
\multirow{-2}{*}{{\color[HTML]{000000} \textbf{\revision{XAI methods}}}} & \multicolumn{1}{c}{{\color[HTML]{000000} }} & \multicolumn{2}{c}{\textbf{\revision{ResNet-101}}}    & \textbf{}     & \multicolumn{2}{r}{\textbf{\revision{ViT-base-16}}}   \\ \cline{1-1} \cline{3-4} \cline{6-7} 
{\color[HTML]{000000} \revision{Occlusion}}                              & \multicolumn{1}{c}{{\color[HTML]{000000} }} & \multicolumn{2}{c}{\revision{0.351 (0.002)}}          &               & \multicolumn{2}{r}{\revision{0.370 (0.003)}}          \\
{\color[HTML]{000000} \revision{RISE}}                                   & \multicolumn{1}{c}{{\color[HTML]{000000} }} & \multicolumn{2}{c}{\revision{0.323 (0.002)}}          &               & \multicolumn{2}{r}{\revision{0.332 (0.002)}}          \\
\revision{SHAP}                                                          &                                             & \multicolumn{2}{c}{\revision{0.341 (0.004)}}          &               & \multicolumn{2}{r}{\revision{0.361 (0.003)}}          \\
\revision{LIME}                                                          &                                             & \multicolumn{2}{c}{\revision{0.335 (0.006)}}          &               & \multicolumn{2}{r}{\revision{0.340 (0.005)}}          \\
\hdashline
\revision{DAX-V1}                                                        &                                             & \multicolumn{2}{c}{\revision{0.318 (0.004)}}          &               & \multicolumn{2}{r}{\revision{0.326 (0.004)}}          \\
\revision{DAX-V2}                                                        &                                             & \multicolumn{2}{c}{\underline{\revision{0.309 (0.004)}}}          &               & \multicolumn{2}{r}{\underline{\revision{0.315 (0.004)}}}          \\
\revision{DAX-V3}                                                        &                                             & \multicolumn{2}{c}{\textbf{\revision{0.298 (0.003)}}} &               & \multicolumn{2}{r}{\textbf{\revision{0.302 (0.002)}}} \\ \hline
\end{tabular}
\egroup
}
\vspace{0.04in}
\caption{\revision{The mean area-under-curve (AUC) for the deletion curves obtained for VOC dataset for using different XAI methods.}}
\label{tab:deletion-auc-voc}
\vspace{-0.2in}
\end{table}
\begin{table}[t!]
\centering
\resizebox{0.9\linewidth}{!}{
\bgroup
\def\arraystretch{1.28}
\begin{tabular}{lllllll}
\toprule[0.75pt]
\multicolumn{1}{c}{{\color[HTML]{000000} \textbf{\revision{DAX variants}}}} &  & \multicolumn{3}{c}{\textbf{\revision{Hyperparameters}}}                & {\color[HTML]{000000} \textbf{}}            & \multicolumn{1}{c}{\textbf{\revision{IoU}}}  \\ \cline{1-1} \cline{3-5} \cline{7-7} 
{\color[HTML]{000000} }                                          &  &                                     &  & \revision{4000, 0.001, 0.01}  & \multicolumn{1}{c}{{\color[HTML]{000000} }} & \multicolumn{1}{c}{\revision{28.5}}          \\
{\color[HTML]{000000} }                                          &  &                                     &  & \revision{4000, 0.001, 0.02}  &                                             & \revision{29.7}                              \\
{\color[HTML]{000000} }                                          &  &                                     &  & \revision{4000, 0.001, 0.04}  &                                             & \revision{29.5}                              \\
{\color[HTML]{000000} }                                          &  &                                     &  & \revision{6000, 0.001, 0.02}  &                                             & \textbf{\revision{31.4}}                     \\
{\color[HTML]{000000} }                                          &  &                                     &  & \revision{6000, 0.01, 0.02}   &                                             & \revision{30.9}                              \\
\multirow{-6}{*}{{\color[HTML]{000000} \revision{V1}}}                      &  & \multirow{-6}{*}{\revision{\{Q, $\lambda_1$, $\lambda_2$\}}} &  & \revision{6000, 0.0001, 0.02} &                                             & \revision{29.2}                              \\ \cline{1-1} \cline{3-5} \cline{7-7} 
{\color[HTML]{000000} }                                          &  &                                     &  & \revision{4000, 0.1}          & \multicolumn{1}{c}{{\color[HTML]{000000} }} & \multicolumn{1}{c}{\revision{28.3}}          \\
{\color[HTML]{000000} }                                          &  &                                     &  & \revision{4000, 0.5}          &                                             & \revision{30.8}                              \\
\multirow{-3}{*}{{\color[HTML]{000000} \revision{V2}}}                      &  & \multirow{-3}{*}{\revision{\{Q, $\lambda_3$\}}}       &  & \revision{6000, 0.5}          &                                             & \textbf{\revision{32.5}}                     \\ \cline{1-1} \cline{3-5} \cline{7-7} 
                                                                 &  &                                     &  & \revision{4000, 0.1}          &                                             & \revision{30.5}                              \\
                                                                 &  &                                     &  & \revision{4000, 0.5}          &                                             & \revision{32.9}                              \\
\multirow{-3}{*}{\revision{V3}}                                             &  & \multirow{-3}{*}{\revision{\{Q, $\lambda_4$\}}}       &  & \revision{6000, 0.5}          &                                             & \multicolumn{1}{c}{\textbf{\revision{34.1}}} \\ \bottomrule[0.75pt]
\end{tabular}
\egroup
}
\vspace{0.04in}
\caption{\revision{Ablation experiments for the hyperparameter choices of DAX variants using the ViT black-box model. }}
\label{tab:ablation-dax-all}
\vspace{-0.2in}
\end{table}
\begin{table}[t!]
\centering
\resizebox{1.0\linewidth}{!}{
\bgroup
\def\arraystretch{1.18}
\begin{tabular}{llllr}
\toprule[0.75pt]
\textbf{\revision{Baselines}}                                                                   &           & \multicolumn{1}{c}{\textbf{\revision{ResNet-101}}} &  & \textbf{\revision{Vit base-16}} \\ \cline{1-1} \cline{3-3} \cline{5-5} 
\begin{tabular}[c]{@{}l@{}}\revision{\textbf{DAX-Base1}: stu. ntw. + RISE}\end{tabular}        &           & \revision{31.6}                                    &  & \revision{31.3}                 \\
\begin{tabular}[c]{@{}l@{}}\revision{\textbf{DAX-Base2}: student network + LIME}\end{tabular}        &           & \revision{29.1}                                    &  & \revision{26.4}                 \\
\textbf{\begin{tabular}[c]{@{}l@{}}\revision{DAX-V2: mask-gen. + stu. ntw.} \end{tabular}} & \textbf{} & \revision{\textbf{34.7}}                                    &  & \revision{\textbf{32.5}}                 \\ \bottomrule[0.75pt]
\end{tabular}
\egroup
}
\vspace{0.03in}
\caption{\revision{IoU comparison between DAX and versions which combine student network distillation and existing XAI methods of LIME/RISE for the  object classification task with ViT black-box.}}
\label{tab:dame-baseline1-compare}
\end{table}
\begin{table}[t!]
\centering
\resizebox{0.95\linewidth}{!}{
\bgroup
\def\arraystretch{1.18}
\begin{tabular}{llllr}
\toprule[0.75pt]
\textbf{\begin{tabular}[c]{@{}l@{}}\revision{DAX student architecture}\end{tabular}}                                  &           & \multicolumn{1}{c}{\textbf{\revision{ResNet-101}}} &  & \textbf{\revision{Vit base-16}} \\ \cline{1-1} \cline{3-3} \cline{5-5} 
\begin{tabular}[c]{@{}l@{}}\revision{2 CNN + 1 FCN layers}\end{tabular}                  &           & \revision{34.8}                                    &  & \revision{31.9}                 \\
\begin{tabular}[c]{@{}l@{}}\revision{2 FCN layers}\end{tabular}                          &           & \revision{35.5}                                    &  & \revision{33.7}                 \\
\textbf{\begin{tabular}[c]{@{}l@{}}\revision{2 CNN + 2 FCN layers}\end{tabular}} & \textbf{} & \textbf{\revision{36.9}}                                    &  & \textbf{\revision{34.1}}                 \\ \bottomrule[0.75pt]
\end{tabular}
\egroup
}
\vspace{0.03in}
\caption{\revision{Architectural choices of student network of DAX}}
\label{tab:arch-choice-dax}
\vspace{-0.15in}
\end{table}

\begin{figure}[t!]
\centering
\includegraphics[width=0.47\textwidth, height=0.28\textwidth]{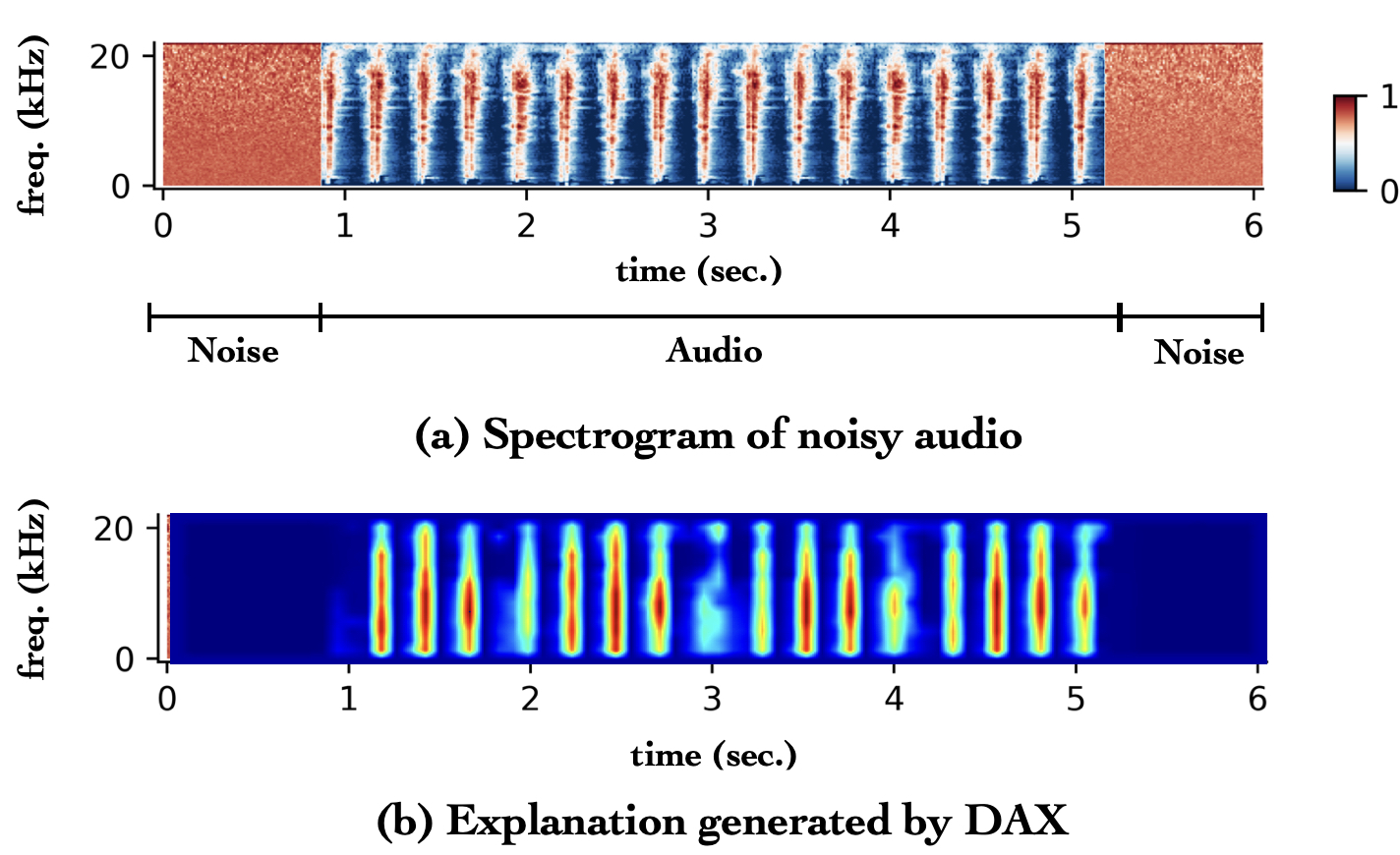}
\caption{The explanation generated by DAX for audio spectogram in the audio classification task. The explanation is not affected by the added noise at both sides of the audio.
}
\vspace{-0.05in}
\label{fig:dame-audio-saliency}
\end{figure}
\revision{\subsubsection{Ablation studies}: For the image processing task, we conducted several ablation studies to understand the importance of different modeling choices made in DAX framework.\\ 
\textbf{Number of perturbation samples and loss weight}: For the different DAX versions, we performed an ablation study with different values of ($Q, \lambda$) and these are reported in Table~\ref{tab:ablation-dax-all}. It is seen that $6000$ perturbation samples, with a $\lambda = 0.5$ provides the best IoU results for both DAX-V2 and DAX-V3 methods. \\
\textbf{Using student distillation with existing XAI approaches}: We explore the option of developing a local distillation using a student model and performing LIME/RISE based explanations on these student models. This is compared with the proposed DAX setting, where both the student and explanation models are jointly optimized. The comparison, shown in Table~\ref{tab:dame-baseline1-compare}, highlights that the joint optimization in DAX setting gives the best IoU results over these other modeling choices.\\
\textbf{Model architecture}: Different configuration of fully connected or CNN architectures are explored for constructing the DAX model. These results are reported in Table~\ref{tab:arch-choice-dax}.}
\begin{table}[t!]
\centering
\resizebox{0.9\linewidth}{!}{
\bgroup
\def\arraystretch{1.3}%
\begin{tabular}{lcr}
\toprule[0.75pt]
\multicolumn{1}{c}{{\color[HTML]{000000} \textbf{XAI Methods}}} & \multicolumn{1}{l}{{\color[HTML]{000000} }} & {\color[HTML]{000000} \textbf{Mean IoU (\%) on ESC-10}} \\ \cline{1-1} \cline{3-3} 
{\color[HTML]{000000} Vanilla gradient~\cite{simonyan2014visualising}}                                    & {\color[HTML]{000000} }                     & {\color[HTML]{000000} 15.5 (8.1)}                       \\
{\color[HTML]{000000} Smooth-grad~\cite{smilkov2017smoothgrad}}                                         & {\color[HTML]{000000} }                     & {\color[HTML]{000000} 23.2 (6.7)}                       \\
{\color[HTML]{000000} Integrated gradients~\cite{sundararajan2017axiomatic}}                                & {\color[HTML]{000000} }                     & {\color[HTML]{000000} 20.1 (6.5)}                       \\
{\color[HTML]{000000} Input x gradient~\cite{shrikumar2017learning}}                                    & {\color[HTML]{000000} }                     & {\color[HTML]{000000} 19.1 (8.7)}                       \\ \hline
{\color[HTML]{000000} GradCAM~\cite{selvaraju2017grad}}                                             & {\color[HTML]{000000} }                     & {\color[HTML]{000000} 24.4 (11.9)}                      \\
{\color[HTML]{000000} Guided GradCAM~\cite{selvaraju2017grad}}                                      & {\color[HTML]{000000} }                     & {\color[HTML]{000000} 19.4 (4.6)}                       \\
{\color[HTML]{000000} GradCAM++~\cite{chattopadhay2018grad}}                                           & {\color[HTML]{000000} }                     & {\color[HTML]{000000} 25.7 (8.2)}                       \\ \hline
{\color[HTML]{000000} Occlusion~\cite{zeiler2014visualizing}}                                           & {\color[HTML]{000000} }                     & {\color[HTML]{000000} 20.1  (5.7)}                      \\
{\color[HTML]{000000} RISE~\cite{petsiuk2018rise}}                                                & {\color[HTML]{000000} }                     & {\color[HTML]{000000} { 21.5 (4.5)}}                 \\
{\color[HTML]{000000} SHAP~\cite{lundberg2017unified}}                                                & {\color[HTML]{000000} }                     & {\color[HTML]{000000} 15.2 (3.3)}                       \\
{\color[HTML]{000000} LIME~\cite{ribeiro2016should}}                                                & {\color[HTML]{000000} }                     & {\color[HTML]{000000} 17.0 (5.3)}                       \\
\hdashline
{\color[HTML]{000000} \textbf{\revision{DAX-V1}}}                               & {\color[HTML]{000000} }                     & {\color[HTML]{000000}  \revision{  22.3 (4.2)}}              \\
{\color[HTML]{000000} \textbf{\revision{DAX-V2}}}                               & {\color[HTML]{000000} }                     & {\color[HTML]{000000} \textbf{\revision{24.1 (4.3)}}}              \\
\bottomrule[0.8pt]
\end{tabular}
\egroup
}
\vspace{0.05in}
\caption{The mean IoU (\%) with associated standard deviation results for various XAI methods on the audio event classification task. The DAX framework  gives the best IoU values among the model agnostic gradient-free methods (last set of rows). 
}
\label{tab:audio-results-iou}
\vspace{-0.2in}
\end{table}
\subsection{Audio processing: sound event classification task}
\noindent \textbf{Task}: We take up an sound event classification task, where the task is to classify an audio signal among $10$ classes.
\\\textbf{Dataset}: We use Environmental Sound Classification (ESC-$10$)~\cite{piczak2015esc} dataset that has $10$ classes of sounds.
\\\textbf{Black-box}: We use a ResNet-$101$ model for this task. The black-box was trained on spectrogram features of audio samples in the training split of ESC-$10$ before performing the explainability analysis on the test set of ESC-$10$, having $150$ audio samples from $10$ classes. The model accuracy was $92.7\%$   and the explainability analysis was performed on correctly predicted $139$ samples from the test set.
\\\textbf{Results}:
Using the  black-box, the explainability analysis was done on the spectrogram of audio samples in the test set of ESC-$10$.  The dataset does not provide time aligned annotations of the sound events. As a way around, we augment the original audio files with noise, of duration ranging from $1$-$5$ sec. at $10$ dB signal-to-noise (SNR) ratio, randomly at both ends of the audio. The original audio segment location forms the ground truth for these explanations, and the task of XAI method is to locate the audio event in the noise-padded   audio file. The IoU values (1-D measure) compute the intersection-over-union of the ground truth audio location (starting to end location of the original audio file) with XAI model's explanation mask.  

As shown in Figure~\ref{fig:dame-audio-saliency}, the DAX output explanations for spectrogram of noisy audio (figure \ref{fig:dame-audio-saliency}(a))   only highlights the audio regions although noise was added (figure \ref{fig:dame-audio-saliency}(b)). Also, the statistical analysis in terms of  the mean IoU values using our approach (DAX) and $11$ other approaches are shown in Table~\ref{tab:audio-results-iou}. Our approach is seen to provide the best mean IoU  value, among all the model-agnostic gradient-free approaches compared for this task.
\begin{table*}[t!]
\centering
\resizebox{0.73\textwidth}{!}{
\def\arraystretch{1.3}%
\begin{tabular}{lccccccr}
\hline
    & \multicolumn{1}{l}{}                        & \multicolumn{6}{l}{}                                                                                                                             \\
\multicolumn{1}{c}{{\color[HTML]{000000} \textbf{XAI Methods}}} & \multicolumn{1}{l}{{\color[HTML]{000000} }} & \multicolumn{6}{c}{{\color[HTML]{000000} \textbf{Mean IoU (\%) on the COSWARA dataset \cite{bhattacharya2023coswara}. }}}                                                             \\ \cline{3-8} 
                                                                           & \multicolumn{1}{l}{}                        & Cough ($\uparrow$)                             & \begin{tabular}[c]{@{}c@{}}Throat  \\ clearing ($\downarrow$)\end{tabular} & Noise  ($\downarrow$)    & Inhale ($\downarrow$)    & Compress ($\downarrow$) & Silence ($\downarrow$)  \\ \hline
{\color[HTML]{000000} Occlusion }                                           & {\color[HTML]{000000} }                     & {\color[HTML]{000000} 32.1 (5.7)} & 26.7 (5.4)                                                 & 22.5 (3.5) & 16.3 (2.5) & 8.1 (4.2) & \textbf{4.1} (1.9) \\
{\color[HTML]{000000} RISE \cite{petsiuk2018rise}}                                                & {\color[HTML]{000000} }                     & {\color[HTML]{000000} 42.3 (4.5)} & 26.5 (5.5)                                                 & 24.6 (4.9) & 13.5 (3.9) & \textbf{6.2} (2.5) & 6.1 (2.2) \\
{\color[HTML]{000000} SHAP \cite{lundberg2017unified}}                                                & {\color[HTML]{000000} }                     & {\color[HTML]{000000} 33.5 (3.3)} & 21.9 (3.8)                                                 & 25.1 (3.1) & 19.3 (4.1) & 7.5 (4.4) & 5.2 (2.4) \\
{\color[HTML]{000000} LIME \cite{ribeiro2016should}}                                                & {\color[HTML]{000000} }                     & {\color[HTML]{000000} 36.4 (6.3)} & \textbf{16.8} (3.4)                                                 & 23.2 (5.8) & 18.3 (3.1) & 7.1 (2.3) & 5.1 (1.9) \\
\hdashline
{\color[HTML]{000000} \textbf{\revision{DAX-V1}}}                               & {\color[HTML]{000000} }                     & {\color[HTML]{000000} \revision{46.4 (5.2)}} & \revision{25.1 (4.3)}                                                 & \textbf{\revision{20.2 (4.1)}} & \revision{{12.3 (4.2)}} & \revision{7.4 (3.5)} & \revision{5.9 (2.1)} \\
{\color[HTML]{000000} \textbf{\revision{DAX-V2}}}                               & {\color[HTML]{000000} }                     & {\color[HTML]{000000} \textbf{\revision{48.1 (4.9)}}} & \revision{23.6 (4.1)}                                                 & \revision{{21.8 (3.6)}} & \textbf{\revision{9.6 (3.8)}} & \revision{8.1 (3.1)} & \revision{4.3 (2.8)} \\
\hline
\end{tabular} 
}
\vspace{0.1in}
\caption{The mean IoU values with standard deviation for the proposed (DAX) framework and other gradient-free model agnostic XAI methods. The ground-truth reference of these audio recordings is obtained from manual annotation \cite{zhu2023importance}. The DAX approach is shown to provide the explanation for the COVID-19 classification by the black-box model primarily based on cough regions.}
\label{tab:coswara-results-iou}
\vspace{-0.2in}
\end{table*}
\subsection{Biomedical: Diagnosing COVID-19 from cough}
Explainability finds one of its prominent requirement in biomedical applications. We study our approach and other existing approaches on the Coswara dataset~\cite{coswara-web},  which contains cough recordings along with models trained to classify COVID and non-COVID participants.
\\\textbf{Task}: The task is the diagnosis of COVID-19 from respiratory sounds like cough sounds collected from COVID-19 positive and healthy individuals, as part of the Coswara project~\cite{coswara-web}.
\\\textbf{Dataset}: The Coswara dataset~\cite{bhattacharya2023coswara} has sound recordings from different categories like breathing, cough, and speech. Our analysis in this work reports on the cough sound based task.  The ground truth segmentation for the different sound events in the cough audio recording is obtained from sound event annotations released  by Zhu et. al.~\cite{zhu2023importance}. This annotation provides time regions corresponding to ``silence'', ``inhalation'', ``compression'', ``noise'', ``throat-clearing'' and ``cough'' events in the given recording.
\\\textbf{Black-box}: The black-box classifier under explainability analysis is the bi-directional LSTM (BLSTM) based baseline classifier provided as part of the diagnosis of COVID-19 using acoustics (DiCOVA) challenge~\cite{sharma2022second}. The model was trained on windowed segments of mel-spectrogram features and provides an  area under receiver-operating-characteristics (ROC) curve   of $79.8$\%. Similar to other tasks, we only analyze the correctly classified samples by the black-box model. A total of $73$ cough samples ($15$ COVID positive and $58$ non-COVID subjects) are evaluated in this analysis.
\\\textbf{Results}: In this analysis, we compare only the gradient-free methods, as shown in Table~\ref{tab:coswara-results-iou}. 
The IoU values obtained for voiced events (cough and throat clearing) are significantly higher than other unvoiced regions, which is consistently shown by all the XAI methods.  Further, the DAX framework is seen to associate the model decisions of COVID/non-COVID more to the ``cough'' regions   than the other baseline systems compared.  {These IoU results, along with the results reported for all the other tasks, consistently show that the proposed DAX framework is capable of explaining the model decisions for diverse tasks with state-of-art choices of black-box models.} 
\begin{figure}[t!]
\centering
\includegraphics[width=0.47\textwidth, height=0.19\textwidth]{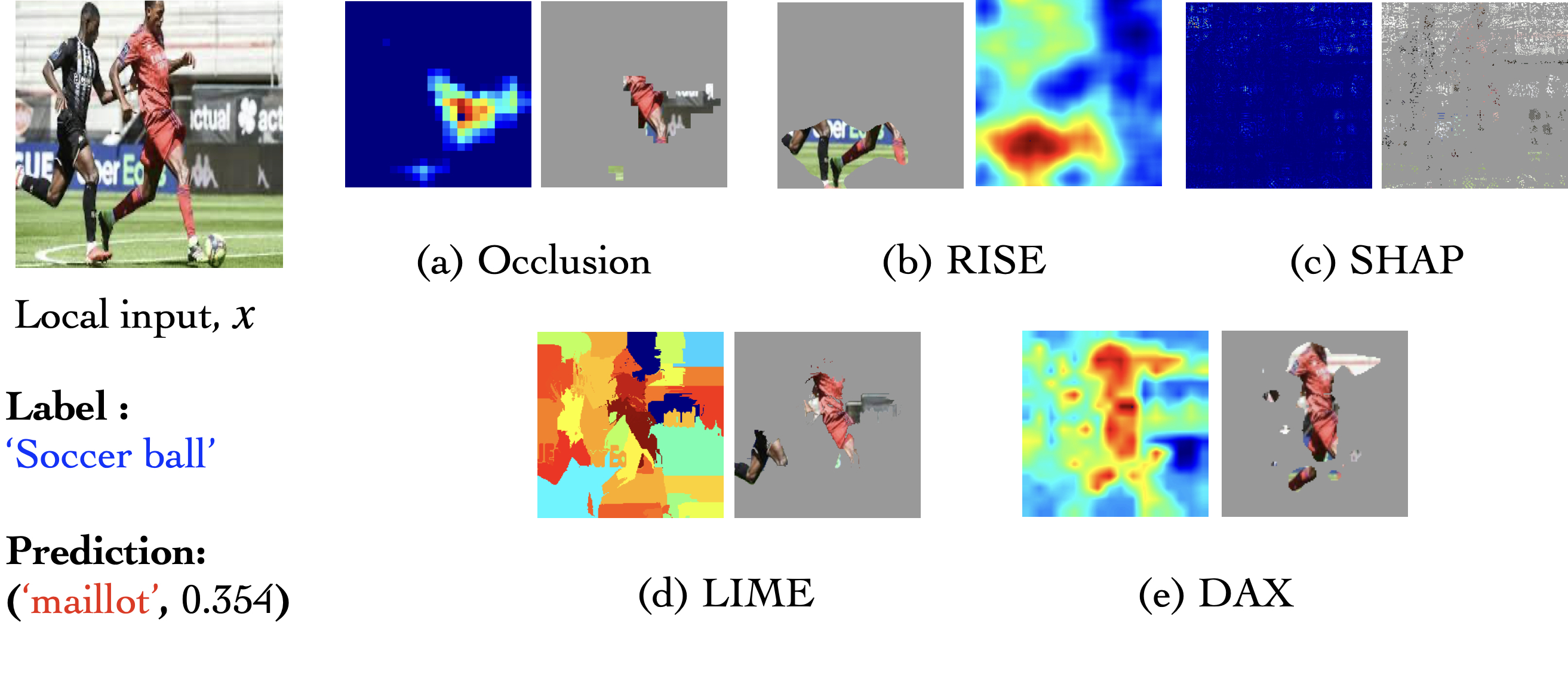}
\vspace{-0.1in}
\caption{Explanations $E[x, T]$ ($T:$ \textit{soccer ball}) generated by different methods for an input sample with true label \textit{soccer ball} which is wrongly predicted as \textit{maillot} by the black-box. As seen, none of the XAI methods show the heat-map for the soccer ball. 
}
\vspace{-0.2in}
\label{fig:negative-sample-saliency-example}
\end{figure}
\begin{table}[t!]
\centering
\resizebox{0.9\linewidth}{!}{
\bgroup
\def\arraystretch{1.25}
\begin{tabular}{lccc}
\toprule[0.75pt]
\multicolumn{1}{c}{{\color[HTML]{000000} \textbf{\begin{tabular}[c]{@{}c@{}}Explainability \\ Methods\end{tabular}}}} & \multicolumn{1}{l}{{\color[HTML]{000000} }} & {\color[HTML]{000000} \textbf{\begin{tabular}[c]{@{}c@{}}Mean IoU \\ (\%) ($\downarrow$) \end{tabular}}} & \multicolumn{1}{c}{\textbf{\begin{tabular}[c]{@{}c@{}}Mean difference from\\  accurate predictions ($\uparrow$) \end{tabular}}} \\ \cline{1-1} \cline{3-4} 
{\color[HTML]{000000} Occlusion~\cite{zeiler2014visualizing}}                                                                                      & {\color[HTML]{000000} }                     & {\color[HTML]{000000} 18.9 (14.5)}                                                              & 9.6                                                                                                         \\
{\color[HTML]{000000} RISE~\cite{petsiuk2018rise}}                                                                                           & {\color[HTML]{000000} }                     & {\color[HTML]{000000} 19.7 (14.2)}                                                              & 11.4                                                                                                        \\
{\color[HTML]{000000} SHAP~\cite{lundberg2017unified}}                                                                                           & {\color[HTML]{000000} }                     & {\color[HTML]{000000} 13.6 (10.0)}                                                              & 5.7                                                                                                         \\
{\color[HTML]{000000} LIME~\cite{ribeiro2016should}}                                                                                           & {\color[HTML]{000000} }                     & {\color[HTML]{000000} 16.6 (13.2)}                                                              & 10.2                                                                                                        \\
\hdashline
{\color[HTML]{000000} \textbf{\revision{DAX-V1}}}                                                                          & {\color[HTML]{000000} }                     & {\color[HTML]{000000} 16.2 (13.9)}                                                         &      15.2                                                                                                      \\ 
{\color[HTML]{000000} \textbf{\revision{DAX-V2}}}                                                                          & {\color[HTML]{000000} }                     & {\color[HTML]{000000} \textbf{\revision{16.0 (13.6)}}}                                                         &      \textbf{\revision{15.4}}                                                                                                      \\
\bottomrule[0.75pt]
\end{tabular}
\egroup
}
\vspace{0.05in}
\caption{Mean IoU values with standard deviation obtained using explanations generated for $141$ incorrectly predicted images using ground truth labels as target class. The IoU for these samples are expected to be lower, which may explain the black-box model's inability to predict the correct class.}
\label{tab:negative-samples-iou}
\vspace{-0.25in} 
\end{table}

\section{Discussion and Conclusion}
\subsection{Explaining Negative Samples}
All the analysis reported thus far only considered samples that were correctly predicted (positive) by the black-box model. In other words, the explanation was sought for positively predicted samples. For several applications, the explanation for the samples which were inaccurately predicted (negative) by the black-box model are also equally important. 
For the Pascal VoC task, we apply the XAI methods on the $141$ samples which are incorrectly predicted by the ViT model. 

In the IoU evaluation, we compare the XAI output with the ground truth segmentation of the target class. Thus, a lower value of IoU indicates that the black-box model prediction was inaccurate as the model was focussing on the less salient regions of the image.
The mean IoU results for explaining these negative samples is reported in Table~\ref{tab:negative-samples-iou}. As seen here, all the models see a degradation in the mean IoU values compared to the ones seen for positive samples (Table~\ref{tab:vision-results-iou}).

We expect the XAI methods to show a large deviation in IoU values for these negative samples from those which are positively classified. The mean IoU difference between positive and negative samples is also shown in this Table. The SHAP model generates the lowest IoU results for these evaluations. However, the mean difference is considerably small, indicating that the SHAP model does not allow a strong distinction between positive and negative samples. Among all the gradient-free post-hoc explainability methods, the DAX framework has the highest difference in IoU values between positive and negative samples, indicating that the XAI method is also applicable for explaining wrong decisions by the black-box model.   
An illustrative example of explaining a negative sample is also shown in Figure~\ref{fig:negative-sample-saliency-example}. 
\subsection{Summary}
\label{sec:conclusion}
This paper proposes an approach for post-hoc gradient-free explainability, called distillation aided explanations (DAX). The DAX framework poses model explanations as a learnable function, given the input sample, its perturbed versions and the corresponding black-box model outputs. Without gradient access, the learning of the DAX function is achieved using a distilled student network that locally approximates the black-box model. Both the masking network and the distillation are jointly optimized to generate the explanation function.

The DAX framework is evaluated on image and audio sub-tasks with state-of-art black box models. The qualitative examples show that the DAX framework generates consistent explanations for wide variety of image classes (Figure  \ref{fig:vision-results-all}). Further, quantitative evaluations are performed by comparing the model explanations with ground truth segments of the objects/audio-events, for the correctly classified samples.  The experiments on diverse audio and image tasks explored in this paper illustrate the benefits of DAX framework in comparison with a variety of other XAI benchmarks.  
\section{Acknowledgement}
\label{sec:ack}
\revision{This work was supported by grants from Verisk Analytics as well as grants from Qualcomm Innovation Fellowship. We are grateful to Maneesh Singh, Chayan Sharma and Rahul Mitra  for the valuable feedback during the conduct of this work.}
\bibliographystyle{IEEEbib}
\bibliography{references}
\end{document}